\documentclass{article} 
\usepackage{iclr2025_conference,times}

\usepackage{amsmath,amsfonts,bm}

\def\eqref#1{equation~\ref{#1}}

\def\1{\bm{1}}

\DeclareMathAlphabet{\mathsfit}{\encodingdefault}{\sfdefault}{m}{sl}
\SetMathAlphabet{\mathsfit}{bold}{\encodingdefault}{\sfdefault}{bx}{n}

\usepackage{newfloat}
\usepackage{listings}
\usepackage{newfloat}
\usepackage{listings}
\usepackage{moreverb}
\usepackage{mathtools}
\usepackage{amssymb}
\usepackage{graphics}
\usepackage{subfigure}
\usepackage{caption}
\usepackage{extarrows}
\usepackage{color}
\usepackage{framed}
\usepackage{bm}
\usepackage{mathrsfs}
\usepackage{mathabx}
\usepackage{multirow}
\usepackage{longtable}
\usepackage{paralist}
\usepackage{indentfirst}
\usepackage{relsize}
\usepackage{extarrows}
\usepackage{lineno,xcolor}
\usepackage{upgreek}
\usepackage{bm}
\usepackage{mwe}
\usepackage{xcolor}
\usepackage{booktabs}
\usepackage[flushleft]{threeparttable}
\usepackage{caption}
\usepackage{booktabs}
\usepackage{rotating} 
\usepackage{float}
\usepackage{nameref}

\usepackage{hyperref}
\usepackage{url}

\title{Dynamic Universal Approximation Theory: The Basic Theory for Transformer-based Large Language Models}

\author{Wei Wang \\
Department of Comp\\
The Hong Kong Polytechnic University\\
\texttt{weiuat.wang@connect.polyu.hk} \\
\And
Qing Li \\
Department of Comp \\
The Hong Kong Polytechnic University \\
\texttt{qing-prof.li@polyu.edu.hk} \\
}

\iclrfinalcopy 
\begin{document}

\maketitle

\begin{abstract}
Language models have become a focal point in the field of artificial intelligence, especially following the emergence of ChatGPT. Large-scale Transformer networks have quickly become the primary method for advancing natural language processing algorithms. These models, built on the Transformer architecture, are capable of engaging in near-human-like interactions and can even assist in guiding human tasks. Despite their impressive capabilities, there remain some critical theoretical questions in the development of large language models (LLMs): What makes the Transformer architecture so effective in driving intelligent language applications, such as translation and coding? What underlies the in-context learning (ICL) capability of LLMs? How does the LoRA approach enhance the fine-tuning of LLMs? What supports the feasibility of pruning in LLMs? To address these key questions and explore the technical strategies within LLMs, we draw on the Universal Approximation Theory (UAT) and then propose the Dynamic Universal Approximation Theory (DUAT) to provide a theoretical foundation, revealing the mechanisms behind these advancements.

\end{abstract}

\maketitle

\section{Introduction}

In recent years, the rapid emergence of LLMs in the field of artificial intelligence has undoubtedly become one of the most notable advancements within the domain. The core allure of these models stems from their extraordinary capabilities in language processing. Language, as a unique crystallization of human intelligence, serves not only as the external reflection of thought but also as the bridge for communication, the cornerstone for the dissemination of knowledge, and the continuation of civilization, profoundly shaping the identity of humans as a unique species. Thus, endowing machines with the ability to understand and generate language marks a significant leap towards the realization of true artificial intelligence. The emergence of models such as the ChatGPT~\cite{Radford2018ImprovingLU, brown2020language, achiam2023gpt}, the Llama~\cite{touvron2023llama}, and the PaLM~\cite{chowdhery2023palm} vividly demonstrates this point.

A distinctive feature of LLMs is their immense parameter size~\cite{achiam2023gpt,touvron2023llama,chowdhery2023palm, chen2021evaluating, zeng2022glm}, often amounting to hundreds of billions or even trillions (for instance, GPT-3's~\cite{brown2020language} 175 billion parameters and PaLM's~\cite{chowdhery2023palm} 540 billion parameters). This vast parameter scale lays the foundation for their exceptional language processing capabilities and enables them to exhibit almost human-like traits, such as ICL ~\cite{brown2020language,dong2022survey}, instruction following~\cite{sanh2021multitask,ouyang2022training,wei2021finetuned}, and multi-step reasoning~\cite{wei2022chain}. Notably, these colossal models are predominantly trained by tech giants like Google and Microsoft using large-scale GPU clusters~\cite{zhao2023survey}, sparking a research fervor on how to efficiently fine-tune them with limited GPU resources. The advent of Lora~\cite{hu2021lora} fine-tuning technology has provided an effective pathway for this, allowing for the fine-tuning of large models under resource constraints without the need to adjust all parameters of the original model comprehensively. Moreover, model pruning techniques~\cite{sun2023simple,ma2023llm} are crucial for deploying large models in resource-constrained environments, aiming to reduce the model size for operation on smaller devices. Today, LLMs possess a diverse range of functionalities, from translation and text summarization to automatic code generation, demonstrating their versatility.

Despite the rapid advancements of LLMs towards higher intelligence and reliability, the theoretical foundations behind these models remain largely unexplored and shrouded in uncertainty. The scientific community is actively seeking to uncover the intrinsic mechanisms behind their powerful capabilities, including the analysis of ICL mechanisms \cite{xie2021explanation, min-etal-2022-rethinking}. In this context, we leverage UAT and then establish the DUAT as the mathematical essence of Transformer-based LLMs and use the perspective of DUAT to elucidate key technologies and phenomena within LLMs, aiming to provide explanations from a new angle. 

Although previous studies have explored the relationship between Transformer \cite{Vaswani2017AttentionIA} and the UAT~\cite{Yun2019AreTU, Yun2020OnCA, Kratsios2021UniversalAU, Alberti2023SumformerUA}, they rely on a series of complex assumptions and derivations, which limits their generalizability and fails to address certain issues in LLMs in a straightforward manner. In contrast, this paper directly demonstrates the connection between multi-layer Transformer and UAT by representing it in a mathematical format. \citet{Cybenko1989ApproximationBS} and Hornik et al.~\cite{Hornik1989MultilayerFN} established the fundamental mathematical form of UAT, which we extend to the DUAT by leveraging the residual structure of Transformers. This form can account for several key issues in LLMs, such as why LLMs are capable of handling both translation and generation tasks simultaneously, and whether LLMs possess memory capacity~\citep{Wang2024SchrodingersML}. Moreover, DUAT is equally applicable to convolutional neural networks (CNNs)~\cite{wang2024universalapproximationtheorybasic}. Our approach not only provides intuitive explanations for some of the challenges within LLMs but also offers theoretical insights that aid in the design of deep learning networks~\cite{wang2024universalapproximationtheoryfoundations}, covering capabilities that previous theories have not addressed. Our contributions are as follows:

\begin{itemize}\vspace{-3pt}
\item We further develop the UAT to DUAT.

\item We prove that the Transformer is the tangible embodiment of DUAT. \vspace{-3pt}

\item We deliver a rigorous scientific explanation of Transformer-based LLMs through the DUAT. \vspace{-3pt}

\item We explain the characteristics of LLMs, such as ICL, instruction following, multi-step reasoning, and the technologies applied within LLMs like Lora, pruning, and LLMs' strong generalization capabilities. \vspace{-3pt}

\end{itemize}

The structure of this paper is as follows: in Section \ref{section:The Universal Approximation Theory}, we first introduce the UAT and propose that to prove Transformers adhere to the DUAT framework, it is necessary to demonstrate that both the linear layer and the multi-head attention mechanism (MHA) can be expressed in the form of matrix-vector multiplication. Then, in Section \ref{section:The Matrix-Vector Method}, we present a general framework for representing various operations in deep learning as matrix-vector multiplication. In Section \ref{section:UAT for Transformer}, we establish that Transformers fall within the scope of DUAT. Our proof strategy involves representing the linear layer and MHA as matrix-vector operations. Sections \ref{Section: Matrix-Vector Method for Linear} and \ref{section:Matrix-Vector Method for Transformer} provide detailed proofs that the linear layer and MHA can indeed be expressed in this form. In Section \ref{section:The UAT Format of Multi-Layer Transformer}, we present the DUAT representation of a multi-layer Transformer network. In Section \ref{Discussion}, we leverage the DUAT framework to theoretically elucidate several fundamental questions (e.g., generalization ability, \ref{section:What makes Transformer so powerful in LLMs?}, and in-context learning capability, \ref{section:What enables LLMs to possess ICL capability?}) and technical aspects (e.g., feasibility of pruning, \ref{section:What justifies the feasibility of pruning LLMs?}, and the effectiveness of the LoRA scheme, \ref{section:How does the LoRA scheme effectively fine-tune LLMs?}) related to large language models (LLMs). Finally, in Section \ref{section:Rethinking LLMs and future study}, we summarize the current challenges of LLMs and explore potential directions for future research.

\section{The Universal Approximation Theory and Matrix-Vector Method}
\label{section:The Universal Approximation Theory and Matrix-Vector Method}

\subsection{The Universal Approximation Theory}
\label{section:The Universal Approximation Theory}

The Universal Approximation Theorem (UAT)~\citep{Cybenko1989ApproximationBS} remains one of the most widely recognized foundational theories in the field of deep learning to date. However, the theorem primarily applies to the simplest forms of neural networks—single-layer or multilayer perceptrons (MLPs)~\citep{Cybenko1989ApproximationBS, popescu2009multilayer}. Due to the significantly increased complexity of Transformer networks, they cannot be expressed mathematically in the same manner as those governed by UAT. Consequently, the extension of UAT to Transformer networks has yet to be established. 

This paper aims to develop the DUAT (an advance of UAT) and explore Transformer networks within this framework to standardize their mathematical representation. Before achieving this, we will briefly revisit the UAT as initially proposed by \citet{Cybenko1989ApproximationBS}. The theorem encompasses a wealth of conclusions and detailed proofs. Although it has been further developed over time, its core mathematical form remains unchanged. Thus, this paper interprets the theory based on the UAT form articulated by \citet{Cybenko1989ApproximationBS}. 

According to Theorem 2 in \citet{Cybenko1989ApproximationBS}, if $\sigma$ is any continuous sigmoid function, then a finite sum of the following form:

\begin{equation}
\begin{aligned}
G(\mathbf{x})=\sum_{j=1}^N \alpha_j \sigma\left(\mathbf{W}_j^{\mathrm{T}} \mathbf{x}+b_j\right)
\end{aligned}
\label{Eq:UAP}
\end{equation}
is dense in $C\left(\mathbf{I}_n\right)$. Here,$\mathbf{x} \in \mathbf{I}_n $, $\mathbf{W}_j \in \mathbb{R}^n$ and $\alpha_j, b \in \mathbb{R}$ are fixed. For any $f \in C\left(\mathbf{I}_n\right)$ and $\varepsilon>0$, there exists a function $G(\mathbf{x})$:
\begin{equation}
\begin{aligned}
|G(\mathbf{x})-f(\mathbf{x})|<\varepsilon
\end{aligned}
\label{eq:UPA bound}
\end{equation}

This implies that, when $N$ is sufficiently large, a neural network can approximate any continuous function on a closed interval. \citet{Hornik1989MultilayerFN} further demonstrates that multilayer feedforward networks also conform to the UAT, capable of approximating arbitrary Borel measurable functions. Observing Eq. \ref{Eq:UAP}, where the function $G(\mathbf{x})$ yields a scalar output in $\mathbb{R}$, the scenario expands naturally when $G(\mathbf{x})$ maps to $\mathbb{R}^m$, requiring the approximation in each dimension. It becomes evident that to accommodate this multidimensional output, a simple extension to Eq. \ref{Eq:UAP} suffices: the transformation matrix $\mathbf{W}_j$ is revised to the space $\mathbb{R}^{n \times m}$, the bias term $b_j$ is recast as a vector in $\mathbb{R}^m$, and $\alpha_j$ is reshaped into a matrix. Nevertheless, in these formulas, the theorem does not straightforwardly apply to Transformer architectures. However, if we succeed in reformulating both the Linear ($\mathbf{x} | \mathbf{W}$) and MHA ($\mathbf{x} | \mathbf{W}$) components into a unified representation, $\mathbf{W}'\mathbf{x}'$, where $\mathbf{W}$ and $\mathbf{x}$ represent the parameters and inputs for each component respectively, and $\mathbf{W}'$ is derived from $\mathbf{W}$ while $\mathbf{x}'$ signifies a column vector derived from $\mathbf{x}$. It is easy to build the connection between multi-layer Transformer networks and UAT. 


\subsection{The Matrix-Vector Method}
\label{section:The Matrix-Vector Method}

\begin{figure}[htbp]
\centering
\includegraphics[width=0.45\textwidth]{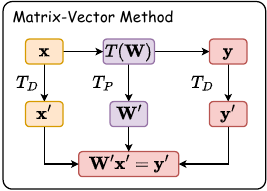}
\caption{The transformation process of the Matrix-Vector Method.}
\label{fig:MVM}
\vspace{-1.5em}
\end{figure}

Before delving into transforming Linear and MHA into their matrix-vector format, we introduce the Matrix-Vector Method, which will subsequently be employed to cast both Linear and MHA operations into a unified matrix-vector format. This method constitutes a strategic realignment of input data and corresponding parameters of various transformations within the network, as illustrated in Figure \ref{fig:MVM}. The underlying principle is as follows: both the input ($\mathbf{x}$) and output ($\mathbf{y}$) data are reconfigured uniformly through a transformation $\mathbf{T_D}$ into column vectors ($\mathbf{x}'$ and $\mathbf{y}'$, while parameter tensor $\mathbf{W}$ is reorganized into matrix form $\mathbf{W}'$ by $\mathbf{T_P}$. $\mathbf{T_D}$ and $\mathbf{T_P}$ are transformations and their purpose is to reorganize the input $\mathbf{x}'$ and $\mathbf{y}'$ into column vectors and $\mathbf{W}'$ fulfill $\mathbf{W}'\mathbf{x}'=\mathbf{y}'$. A critical requirement is that the input and output throughout the LLMs network adhere to the same restructuring scheme, thereby eliminating the need for additional transformations on intermediate feature data; they can directly be represented as column vectors. Additionally, by default, matrix variables in the original formulas are represented in bold, such as $\mathbf{x}$, and in the Matrix-Vector form, corresponding variables are denoted with a prime symbol ($'$) in the upper right corner, such as $\mathbf{x}'$. Elements within matrices are represented by corresponding lowercase letters with subscripts, for example, $x_i$. The Matrix-Vector Method can be succinctly encapsulated as follows:

\begin{equation}
\begin{aligned}
\mathbf{y}=T(\mathbf{x}|\mathbf{W}) \mapsto \mathbf{y}'=\mathbf{W}'\mathbf{x}'
\end{aligned}
\label{eq-Matrix-Vector}
\end{equation}

where $T$ represents a general transformation, like MHA. It is obvious that $T_D$ and $T_P$ are not fixed, we could design various kinds of ways to do those. For convenience, we present a methodology tailored for Linear in Section \ref{Section: Matrix-Vector Method for Linear} and an approach for MHA in Section \ref{section:Matrix-Vector Method for Transformer}, thereby illustrating the adaptability and application of the Matrix-Vector Method across different components of the Transformer architecture.

\begin{figure}[htbp]
\centering
\includegraphics[width=0.8\textwidth]{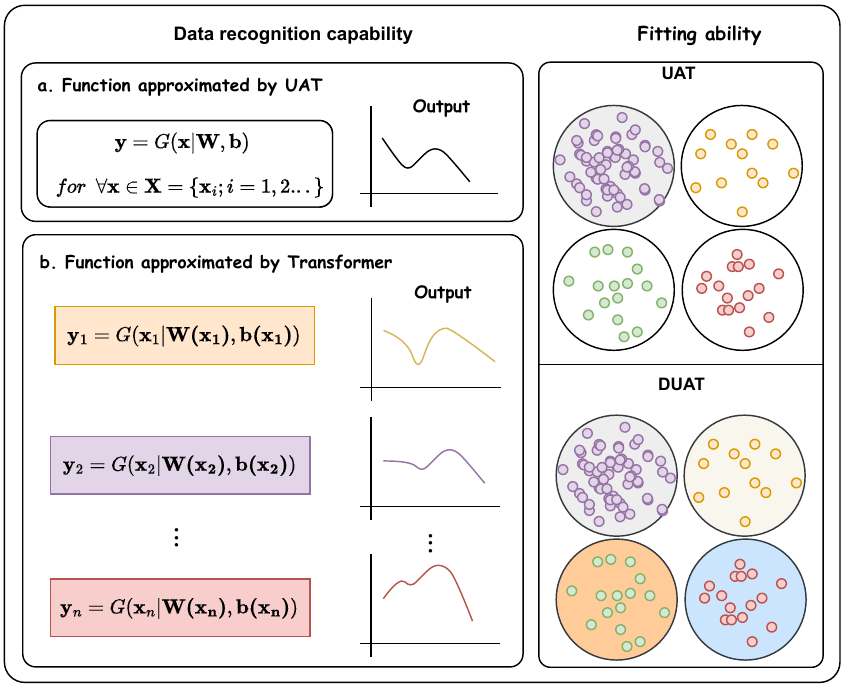}
\caption{
This diagram shows the differences between an UAT and a DUAT. 
} 
\label{fig:Func_Approx}
\vspace{-1.5em}
\end{figure}

\subsection{The introduction of DUAT}
Although the UAT function has strong approximation capabilities, it has a fatal flaw: it can only fit one function within a closed interval. This limitation results in poor generalization.Therefore, to enable the model to fit different functions within a closed interval based on the input, we need to grant the parameters in the UAT the ability to dynamically adjust according to the input, as defined in Eq. \ref{eq:def-DUAT}.

\begin{equation}
\begin{aligned}
\mathbf{y}_i={G(\mathbf{x}_i|\mathbf{W(\mathbf{x}_i)},\mathbf{b(\mathbf{x}_i)})}
\end{aligned}
\label{eq:def-DUAT}
\end{equation}

where $\mathbf{W}=\mathbf{W}_1\cdots \mathbf{W}_N$ and $\mathbf{b}=b_1\cdots b_N$Figure \ref{fig:Func_Approx} illustrates the comparison between UAT and DUAT. On the left side, Figure \ref{fig:Func_Approx}.\textbf{a} illustrates the limitation of the Universal Approximation Theorem (UAT) in fitting only a single function due to its fixed parameters post-training. In contrast, Figure \ref{fig:Func_Approx}.\textbf{b} demonstrates the capability of the Dynamic Universal Approximation Theorem (DUAT) to approximate multiple functions. In DUAT, some or all parameters are functions of the input, allowing it to dynamically adjust the parameters that would be fixed in UAT, thereby fitting different functions based on varying inputs. This adaptability enables DUAT to generate a diverse range of outputs.

On the right side, Figure \ref{fig:Func_Approx} compares the data distribution fitting capabilities of UAT and DUAT. Since UAT has fixed parameters, it can only fit a single, fixed data distribution. Given that training typically involves large datasets, the objective for UAT is to fit as much data as possible, which often results in fitting only the densest regions of the data distribution, highlighted in purple. Conversely, DUAT can dynamically adjust its parameters based on the input, allowing it to simultaneously fit multiple distinct distributions.

This comparison underscores the advantages of DUAT in handling more complex and diverse data scenarios, making it a more versatile tool for various applications.

\section{DUAT for Transformer}
\label{section:UAT for Transformer}

In this section, we will apply the matrix-vector approach to convert the operations in the Transformer \cite{Vaswani2017AttentionIA} into matrix-vector multiplication form, and then explain the defination of DUAT and demonstrate that the Transformer is essentially a DUAT function.

\subsection{Matrix-Vector Method for Linear}
\label{Section: Matrix-Vector Method for Linear}

\begin{figure}[htbp]
\centering
\includegraphics[width=0.5\textwidth]{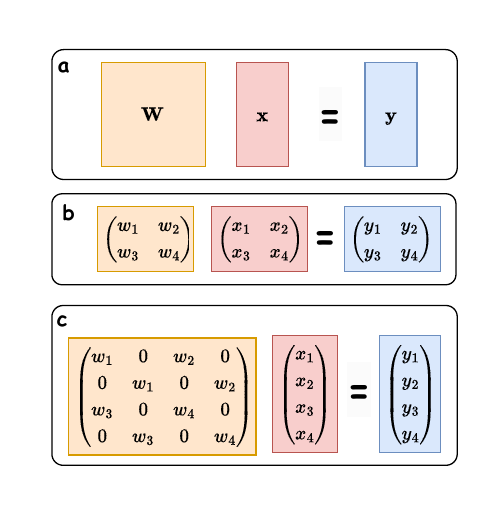}
\caption{The process of converting a linear transformation into its corresponding matrix-vector representation. \textbf{a}: Depicts the general form of a linear transformation.
\textbf{b}: Presents a straightforward example of a linear transformation. \textbf{c}: Demonstrates the transformation of the linear operation from \textbf{b} into the matrix-vector format.}
\label{fig:Linear-MV}
\end{figure}

In this section, we present a way to transform Linear operation into matrix-vector format. Figure \ref{fig:Linear-MV} illustrates this process: Figure \ref{fig:Linear-MV}.\textbf{a} shows the linear transformation of multi-channel input: $\mathbf{W}\mathbf{x}=\mathbf{y}$. Figure \ref{fig:Linear-MV}.\textbf{b} provides a specific example of Figure \ref{fig:Linear-MV}.\textbf{a}, Figure \ref{fig:Linear-MV}.\textbf{c} converts the linear transformation in Figure \ref{fig:Linear-MV}.\textbf{b} into the corresponding matrix-vector representation: $\mathbf{W}'\mathbf{x}'=\mathbf{y}'$. Thus, the linear transformation can be represented in matrix-vector form as follows:

\begin{equation}
\begin{aligned}
\mathbf{x}_{i+1} = \mathbf{W}_i\mathbf{x}_i   \mapsto  \mathbf{x}_{i+1}'=\mathbf{W}'_i\mathbf{x}'_i
\end{aligned}
\label{eq:linear-MV}
\end{equation}

Here, $\mathbf{x}_i\in \mathbb{R}^{(N,M)}$ and $\mathbf{x}_{i+1}\in \mathbb{R}^{(N,M)}$ represent the input and output of layer $i$, respectively, while $\mathbf{W}_i\in \mathbb{R}^{(N,N)}$ represents the parameters of layer $i$. $\mathbf{x}_i'$, $\mathbf{x}_{i+1}'$, and $\mathbf{W}_i'$ are generated based on $\mathbf{x}^i$, $\mathbf{x}^{i+1}$, and $\mathbf{W}_i$ using the Matrix-Vector Method. For convenience, we use $\mathbf{W}_i'$ to represent $(\mathbf{W}_i')^T$. Therefore, $\mathbf{x}_{i+1}' = \mathbf{W}_i'\mathbf{x}_i' $. So the  FFN in the Transformer could be written as:

\begin{equation}
\begin{aligned}
FFN(\mathbf{x}) &= \mathbf{W}_{2}\sigma (\mathbf{W}_{1}\mathbf{x}+\mathbf{b}_{1})+\mathbf{b}_{2}\\
& \mapsto  \mathbf{W}_{2}'\sigma (\mathbf{W}_{1}'\mathbf{x}'+\mathbf{b}_{1}')+\mathbf{b}_{2}'
\end{aligned}
\label{eq:ffn}
\end{equation}

\subsection{Matrix-Vector Method for MHA}
\label{section:Matrix-Vector Method for Transformer}

We now employ the Matrix-Vector Method to elucidate the inner workings of the MHA. The mechanism is defined by the following equation:

\begin{equation}
\begin{aligned}
&\mathring{\mathbf{H}}= \operatorname{MultiHead}(\mathbf{Q}, \mathbf{K}, \mathbf{V}) \\
=&\operatorname{Concat}\left(\hat{\mathbf{H}}_1, \ldots, \hat{\mathbf{H}}_{h}\right) \mathbf{W}_O=\hat{\mathbf{H}}\mathbf{W}_O \\
\end{aligned}
\label{eq:mutil-head-attention}
\end{equation}

\begin{equation}
\begin{aligned}
&\operatorname{Attention}(\mathbf{x}_i \mathbf{W}_{Qi}=\mathbf{Q}_i, \mathbf{x}_i \mathbf{W}_{Ki}=\mathbf{K}_i, \mathbf{x}_i\mathbf{W}_{Vi}=\mathbf{V}_i)\\
&=\operatorname{softmax}\left(\frac{\mathbf{Q}_i \mathbf{K}_i^T}{\sqrt{M}}\right) \mathbf{V}_i\\
&=\mathbf{H}_i\mathbf{V}_i =\mathbf{H}_i[\mathbf{x}_i\mathbf{W}_{Vi}]=\hat{\mathbf{H}}_i \\  
\end{aligned}
\label{eq:mutil-head}
\end{equation}

\begin{figure}[htbp]
\centering
\includegraphics[width=0.75\textwidth]{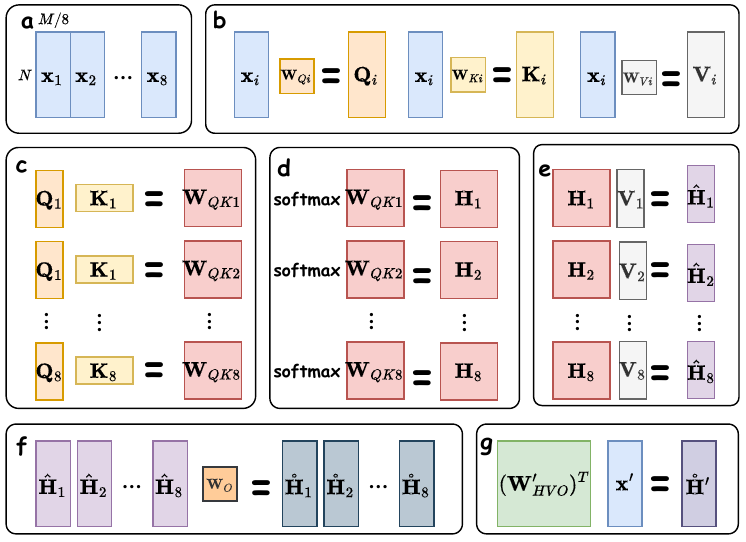}
\caption{The Transformer process. }
\label{fig:TF2M_ALL}
\end{figure}

Here, $h$ represents the number of attention heads, and the input $\mathbf{x}\in \mathbb{R}^{(N,M)}$ is divided into $\mathbf{x}_1,...,\mathbf{x}_h$ based on $h$. The parameters $\mathbf{W}_{Qi}, \mathbf{W}_{Ki}$, and $\mathbf{W}_{Vi}$ correspond to $\mathbf{x}_i$. The whole process of MHA can be represented in Figure~\ref{fig:TF2M_ALL}. Figure~\ref{fig:TF2M_ALL}.\textbf{a} represents that the input $\mathbf{x}$ is split into $\mathbf{x}_1 \cdots \mathbf{x}_8$ based on the number of heads. Figure~\ref{fig:TF2M_ALL}.\textbf{b} represents $\mathbf{x}_i \mathbf{W}_{Qi}=\mathbf{Q}_i, \mathbf{x}_i \mathbf{W}_{Ki}=\mathbf{K}_i, \mathbf{x}_i \mathbf{W}_{Vi}=\mathbf{V}_i$. Figure~\ref{fig:TF2M_ALL}.~\textbf{c}, \textbf{d}, \textbf{e} represent the process of $\operatorname{softmax}\left(\frac{\mathbf{Q}_i \mathbf{K}_i^T}{\sqrt{M}}\right) \mathbf{V}_i$ and we omit $\sqrt{M}$. Figure~\ref{fig:TF2M_ALL}.\textbf{f} represents $\operatorname{Concat}\left(\hat{\mathbf{H}}_1, \ldots, \hat{\mathbf{H}}_{h}\right) \mathbf{W}_O$. Figure~\ref{fig:TF2M_ALL}.\textbf{g} represents the whole process by a matrix multiplication of $(\mathbf{W}'_{HVO})^T\mathbf{x}'=\mathring{\mathbf{H}}'$, where $\mathbf{W}'_{HVO}$ is generated based $\mathbf{H}_1\cdots\mathbf{H}_8$, $\mathbf{W}_{V1}\cdots\mathbf{W}_{V8}$ and $\mathbf{W}_{O}$. This means that we could represent the whole complex MHA in matrix multiplication. Next, we give the proof of this process.

Our objective is to express the MHA as $(\mathbf{W}'_{HVO})^T\mathbf{x}'=\mathring{\mathbf{H}}'$. Prior to transforming the MHA into the matrix-vector form, we need to conduct a comprehensive analysis and clearly define the research object. In Eq.~\ref{eq:mutil-head-attention}, $\hat{\mathbf{H}}=\operatorname{Concat}\left(\hat{\mathbf{H}}_1, \ldots, \hat{\mathbf{H}}_{h}\right)$ describes an engineering process that requires mathematical representation. The learning process for the input $\mathbf{x}$ primarily consists of two parts: $\mathbf{V}_1, \ldots, \mathbf{V}_h$ and $\mathbf{H}_1, \ldots, \mathbf{H}_h$. $\mathbf{H}_i$ is derived based on the parameters $\mathbf{x}_{i}$, $\mathbf{W}_{Qi}$ and $\mathbf{W}_{Ki}$. 

\begin{figure}[htbp]
\centering
\includegraphics[width=0.75\textwidth]{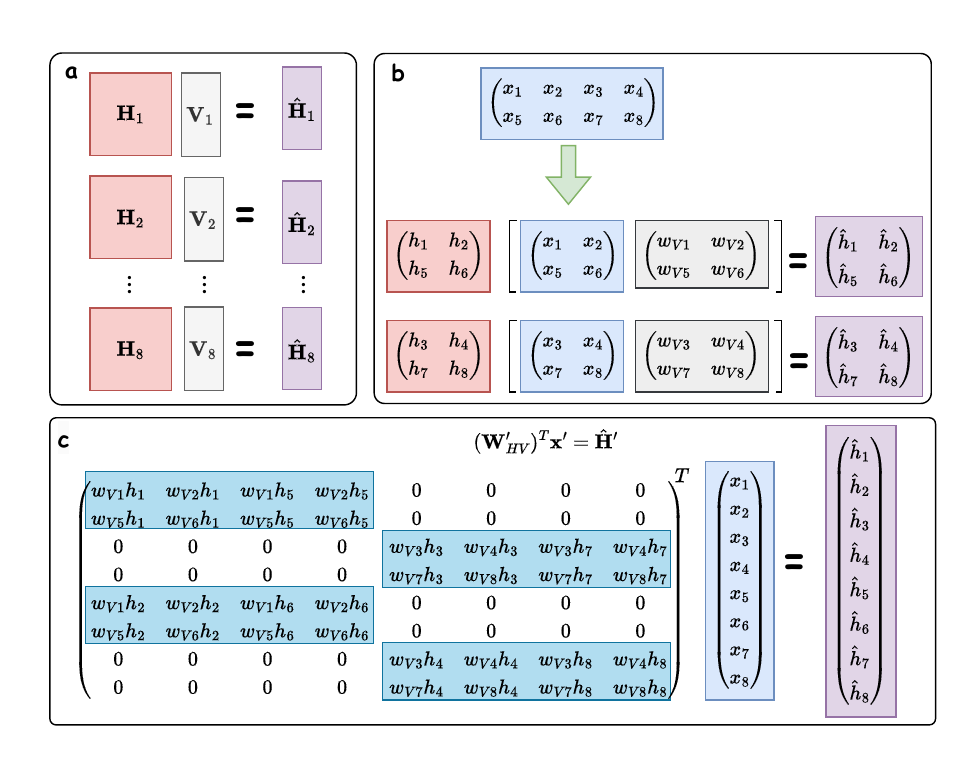}
\caption{The process of transforming $\hat{\mathbf{H}}=\text{Concat}(\hat{\mathbf{H}}_1...\hat{\mathbf{H}}_8)$ in the MHA into its corresponding matrix-vector form $(\mathbf{W}_{HV}')^T\mathbf{x}'=\hat{\mathbf{H}}'$.}
\label{fig:M-Attention}
\vspace{-1.5em}
\end{figure}

Figure \ref{fig:M-Attention}.\textbf{a} represents the computation process of  $\mathbf{H}_1\mathbf{V}_1=\hat{\mathbf{H}}_1, \ldots, \mathbf{H}_{h}\mathbf{V}_h=\hat{\mathbf{H}}_h$. In Figure \ref{fig:M-Attention}.\textbf{b}, we present a simple example of Figure \ref{fig:M-Attention}.\textbf{a}. Figure \ref{fig:M-Attention}.\textbf{b} depicts $\mathbf{H}_i[\mathbf{x}_i\mathbf{W}_{Vi}], i=1,2$. In Figure \ref{fig:M-Attention}.\textbf{c} we convert Figure \ref{fig:M-Attention}.\textbf{b} into the matrix-vector form $(\mathbf{W}_{HV}')^T\mathbf{x}'=(\hat{\mathbf{H}}')^T$, where $\mathbf{W}_{HV}'$ is generated from $\mathbf{H}_{1}, \mathbf{H}_{2}$ and $\mathbf{W}_{V1}, \mathbf{W}_{V2}$. More details can be found in \textcolor{blue}{Appendix B}.

\begin{figure*}[htbp]
\centering
\includegraphics[width=0.85\textwidth]{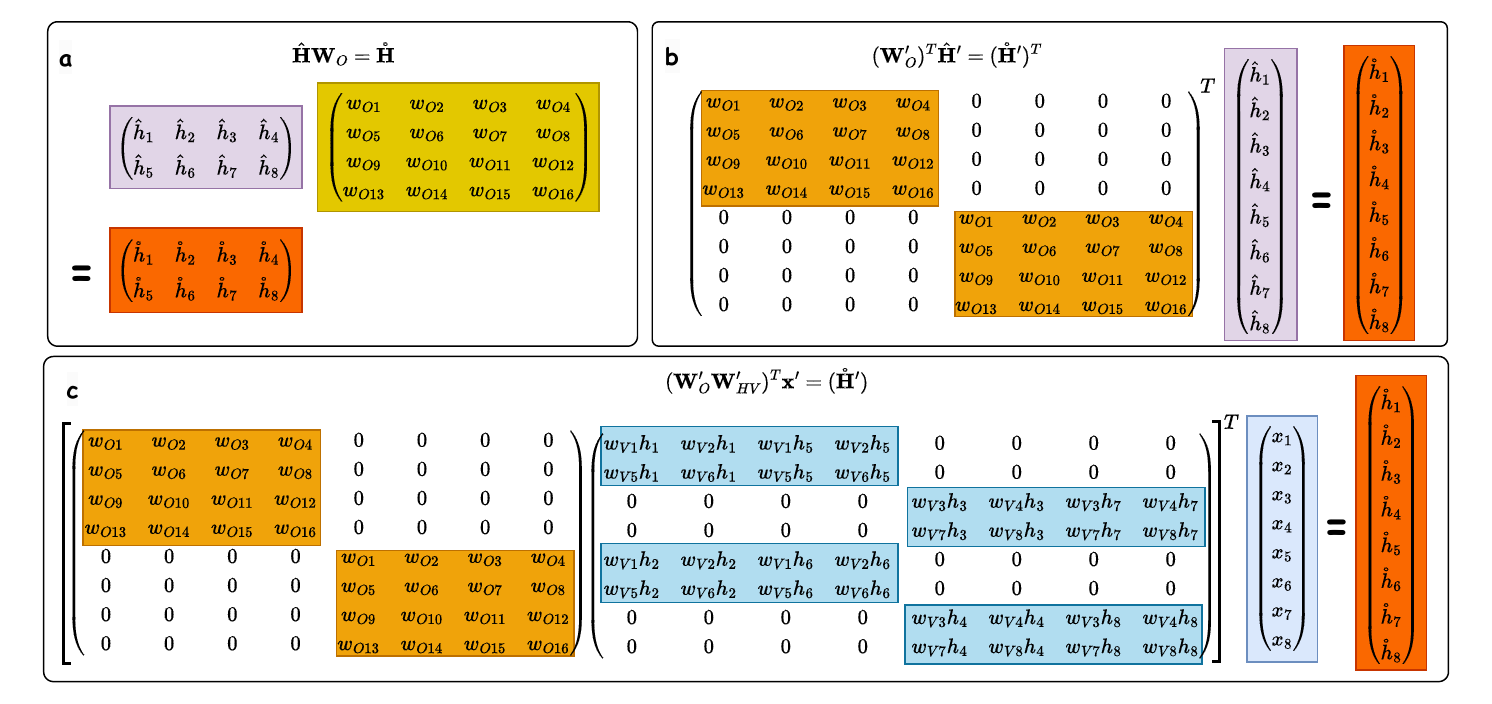}
\caption{This diagram illustrates the process of transforming $\text{Concat}(\hat{\mathbf{H}}_1...\hat{\mathbf{H}})\mathbf{W}_O$ in the MHA into its corresponding matrix-vector form $(\mathbf{W}_{HVO}')^T\mathbf{x}'=\mathring{\mathbf{H}}'$, different colors in the diagram correspond to different variables.}
\label{fig:TF2M_O}
\vspace{-1.5em}
\end{figure*}

Figure \ref{fig:TF2M_O} illustrates the parameter transformation scenario after incorporating $\mathbf{W}_O$. Figure \ref{fig:TF2M_O}.\textbf{a} demonstrates an example of $\mathring{\mathbf{H}}=\operatorname{Concat}\left(\hat{\mathbf{H}}_1, \ldots, \hat{\mathbf{H}}_{h}\right) \mathbf{W}_O$, while Figure \ref{fig:TF2M_O}.\textbf{b} rewrites Figure \ref{fig:TF2M_O}.\textbf{a} as $(\mathbf{W}_O')^T\hat{\mathbf{H}}'=\mathring{\mathbf{H}}'$. Consequently, based on Figures \ref{fig:M-Attention} and \ref{fig:TF2M_O}, the entire MHA can be expressed as $(\mathbf{W}_{HVO}')^T\mathbf{x}'=\mathring{\mathbf{H}}'$, where $\mathbf{W}_{HVO}'=\mathbf{W}_{O}'\mathbf{W}_{HV}'$ (Based on Figures \ref{fig:TF2M_O}, we can get $\mathbf{W}_{HVO}'$, but the resulting expression is too complex and not easy to understand, so we did not compute the final result. However, it is easy to analyze that as long as the variables in $\mathbf{W}_{HV}$ and $\mathbf{W}_{O}'$ are non-zero, the result is essentially a dense matrix). So the whole MHA can be written as:
\begin{equation}
\begin{aligned}
\mathring{\mathbf{H}}'&=(\mathbf{W}_{HVO}')^T\mathbf{x}'\\
\end{aligned}
\end{equation}

Here, $\mathbf{x} \in \mathbb{R}^{(NM,1)}$ and $\mathring{\mathbf{H}} \in \mathbb{R}^{(NM,1)}$ are the input and output, respectively, while $\mathbf{W}_{HVO}' \in \mathbb{R}^{(NM,NM)}$ denotes matrices generated in accordance with $\mathbf{H}$, $\mathbf{W}_V$ and $\mathbf{W}_O$. In this manner, we have expressed MHA as a matrix-vector multiplication. This matrix multiplication representation provides a more concise way to express the MHA mechanism. 

For convenience, we use $\mathbf{W}'$ to represent $(\mathbf{W}_{HVO}')^T$. The MHA operation could be simply written as:

\begin{equation}
\begin{aligned}
\mathring{\mathbf{H}}'=\mathbf{W}'\mathbf{x}'
\end{aligned}
\label{eq:MHA}
\end{equation}

\subsection{The DUAT Format of Multi-Layer Transformer}
\label{section:The UAT Format of Multi-Layer Transformer}

The Transformer architecture is founded on two pivotal components: FFN and MHA. In Sections \ref{Section: Matrix-Vector Method for Linear} and \ref{section:Matrix-Vector Method for Transformer}, we have showcased the matrix-vector representations for both FFN and MHA. In this section, we delve into why multi-layer Transformer is the implementation of DUAT.

Based on Eq \ref{eq:ffn} and \ref{eq:MHA}, we can demonstrate that $i+1$-th layer ($i=0,1,2,\ldots$) Transformer can be written as (see \textcolor{blue}{Appendix C} for more details):

\begin{equation}
\begin{aligned}
\mathbf{x}_{i+1}=&(\mathbf{W}_{i+1,1}'\mathbf{x}_0+\mathbf{b}_{i+1,1})\\
&+\sum_{j=1}^{i+1}\mathbf{W}'_{j,3}\sigma (\mathbf{W}'_{j,2}\mathbf{x}'_{0}+\mathbf{b}'_{j,2})
\end{aligned}
\label{eq:TF}
\end{equation}

where $ i = 0 $, the parameters are defined as follows: $ \mathbf{W}'_{1,1} = \mathbf{W}'_{1,1} $, $ \mathbf{b}'_{1,3} = \mathbf{b}'_{1,3} $, $ \mathbf{W}'_{1,3} = \mathbf{W}'_{1,3} $, $ \mathbf{W}'_{1,2} = \mathbf{W}'_{1,2} \mathbf{W}'_{1,1} $, and $ \mathbf{b}'_{1,2} = \mathbf{b}'_{1,2} $. We define the parameters for cases where $ i \geq 1 $. For each $ j = 1, 2, \ldots, i $, the updates are as follows: $ \mathbf{W}'_{j+1,1} = \mathbf{W}'_{j+1,1} \mathbf{W}_{j,1} $, $ \mathbf{b}'_{j+1,3} = \mathbf{W}'_{j+1,1} \mathbf{b}_{j,3} + \mathbf{b}'_{j+1,3} $, $ \mathbf{W}'_{j+1,2} = \mathbf{W}'_{j+1,2} \mathbf{W}_{j,1}$, and $ \mathbf{W}'_{j,3} = \mathbf{W}'_{j+1,1} \mathbf{W}'_{j,3}$. Additionally, for $ j = 2, \ldots, i + 1 $, the bias terms $ \mathbf{b}'_{j,2} $ are updated according to:
\begin{equation}
\begin{aligned}
\mathbf{b}'_{j,2} =& (\mathbf{W}'_{j,2} \mathbf{b}'_{j-1,3} + \mathbf{b}'_{j,2}) \\
+& \mathbf{W}'_{j,2} \sum_{k=1}^{j-1} \mathbf{W}'_{k,3} \sigma (\mathbf{W}'_{k,2} \mathbf{x}'_{0} + \mathbf{b}'_{k,2}).
\end{aligned}
\label{eq:b_j2}
\end{equation}

The term $\mathbf{b}'_{j,2}$ is approximated by the $j$ layer UAT with $\mathbf{x}_0$ as input. In the MHA mechanism, the parameters change dynamically with the input. Therefore, in the formula above, all $\mathbf{W}'_{j,1}$ and $\mathbf{W}'_{j,2}$ for $i+1 \geq j \geq 1$, and $\mathbf{W}'_{j,3}$ for $i+1 > j > 1$ in layer $i$, are dynamically adjusted based on the input. According to Eq. \ref{eq:TF}, we know the mathematical format of a multi-layer Transformer is the same as UAT, but the parameters in the multi-layer Transformer's UAT are not fixed, they will change with input. So we call the UAT format of parameters dynamically adjust based on input as DUAT.

\section{Discussion}
\label{Discussion}
Leveraging on our proof in the previous section that the Transformer is the tangible embodiment of DUAT, in this section we address the following critical problems and explore the technical strategies of LLMs: Why does the Transformer architecture possess such power in enabling intelligent language models, such as translation and programming? What enables LLMs' capacity for ICL? How does the LoRA scheme effectively fine-tune LLMs? What justifies the feasibility of pruning LLMs?

\subsection{What Makes Transformer So Powerful in LLMs?}\label{section:What makes Transformer so powerful in LLMs?}

Theoretically, we have established that Transformer networks are concrete implementations of the DUAT, enabling them more powerful approximation ability than UAT. While UAT provides powerful function approximation capabilities, it inherently lacks the ability to approximate multiple functions simultaneously. However, language tasks are inherently diverse, often requiring the approximation of different functions based on the input. For instance, when summarizing, translating, or continuing the same text, the input functions are nearly identical, with only minor variations in the prompt. Without the ability to dynamically approximate functions based on the input, simply fitting a general function trend based on input will result in identical or similar outputs.

To address this, LLMs must distinguish and adapt to these nearly identical functions, dynamically generating response functions based on the input. The MHA and residual mechanisms in Transformers equip LLMs with the ability to dynamically approximate relevant functions according to the input. Specifically, MHA allows for the dynamic adjustment of the weight parameters in UAT in response to the input, while the residual mechanism supports the dynamic approximation of bias terms. This theoretical foundation enables Transformer-based LLMs to handle a wide range of tasks, including translation, continuation, summarization, code generation, and solving mathematical problems.

Furthermore, the MHA mechanism can capture global information (as shown in Figure \ref{fig:TF2M_O}, where each element in the output $\mathring{\mathbf{H}}'$ contains global information), which helps generate content that is consistent with the context. This is crucial for language understanding that requires extensive contextual information. In contrast, 1D convolution employs a sliding convolution learning strategy, where the learned content is somewhat influenced by the size of the convolution kernel, leading to a more fragmented learning approach (for more information, please refer to \textcolor{blue}{Appendix D}).

\subsection{What Enables LLMs to Possess ICL Capability?}
\label{section:What enables LLMs to possess ICL capability?}
Contextual interaction, as the core capability of LLMs, permeates every phase from training and fine-tuning to prediction. ICL, multi-step reasoning, and instruction following are intuitive manifestations of this contextual interaction. Leveraging their context-sensitive interaction capabilities, LLMs can exhibit behaviors consistent with ICL, multi-step inference, and instruction following, which are tailored based on contextual cues. 

So, how does this contextual interaction capability arise within LLMs? The formula $\mathring{\mathbf{H}} = (\mathbf{W}'_{HVO})^T\mathbf{x}'$ in Figure \ref{fig:TF2M_O} reveals this mode of contextual interaction. Since $\mathbf{W}'_{HVO}$ represents a dense matrix (almost devoid of zero elements and whose internal elements are highly correlated), each element in $\mathring{\mathbf{H}}$ encapsulates comprehensive information from both preceding and subsequent contexts. This learning of holistic contextual information constitutes the foundation of contextual interaction within LLMs. (See \textcolor{blue}{Appendix D} for more details)

\subsection{What Justifies the Feasibility of Pruning LLMs? }\label{section:What justifies the feasibility of pruning LLMs?}

Due to the massive size of parameters in LLMs and the subsequent high demand for computational resources, pruning LLMs is pivotal for their deployment. A legitimate question to ask is why LLMs are amenable to pruning. The rationale lies in the presence of excessively low-weight parameters in certain layers of LLMs. To understand this, we can directly analyze it from the perspective of the formula underlying the UAT:

\begin{equation} 
\begin{aligned} 
|\sum_{j=1}^N \alpha_j \sigma\left(\mathbf{W}_j^{\mathrm{T}} \mathbf{x}+b_j\right)-f(\mathbf{x})|<\varepsilon \quad  . \end{aligned} 
\end{equation}

for all $\mathbf{x} \in \mathbf{I}_n$. Let's assume $\Lambda=\{1,2...N\},\Lambda_1\cup \Lambda_2=\Lambda,\Lambda_1\cap  \Lambda_2= \emptyset$ and $|\sum_{j\in \Lambda_2} \alpha_j \sigma\left(\mathbf{W}_j^{\mathrm{T}} \mathbf{x}+\theta_j\right)| \to 0$. Then we have:

\begin{equation} 
\begin{aligned} 
&|\sum_{j\in \Lambda_1} \alpha_j \sigma\left(\mathbf{W}j^{\mathrm{T}} \mathbf{x}+b_j\right)\\
+&\sum_{j \in \Lambda_2} \alpha_j \sigma\left(\mathbf{W}_j^{\mathrm{T}} \mathbf{x}+b_j\right)-f(\mathbf{x})|<\varepsilon 
\end{aligned} 
\end{equation}

Since $|\sum_{j\in \Lambda_2} \alpha_j \sigma\left(\mathbf{W}_j^{\mathrm{T}} \mathbf{x}+\theta_j\right)| \to 0$, we have the following inequality:

\begin{equation} 
\begin{aligned}
& \big|| \sum_{j \in \Lambda_1} \alpha_j \sigma\left(\mathbf{W}_j^{\mathrm{T}} \mathbf{x}+\theta_j\right)-f(\mathbf{x}) \mid \\
 -&|\sum_{j \in \Lambda_2} \alpha_j \sigma\left(\mathbf{W} j^{\mathrm{T}} \mathbf{x}+\theta_j\right)| \big|<\varepsilon
\end{aligned}
\end{equation}

Therefore, we have:

\begin{equation} 
\begin{aligned} 
& | \sum_{j \in \Lambda_1} \alpha_j \sigma\left(\mathbf{W}_j^{\mathrm{T}} \mathbf{x}+b_j\right)-f(\mathbf{x}) \mid \\
<&\varepsilon+|\sum_{j \in \Lambda_2} \alpha_j \sigma\left(\mathbf{W} j^{\mathrm{T}} \mathbf{x}+b_j\right)| 
\end{aligned} 
\end{equation}

Hence, when parameters in certain layers are small enough, we can directly remove those layers since their impact on the final result is minimal.

\subsection{How Does the LoRA Scheme Effectively Fine-tune LLMs? }\label{section:How does the LoRA scheme effectively fine-tune LLMs?}

Given the substantial computational resources required to train LLMs and their powerful generalization abilities, we believe that more efficient use of pre-trained models is essential. Re-training models from scratch incurs significant computational costs, so reusing well-trained models for new tasks is both practical and resource-efficient. A prominent solution to this challenge is the LoRA~\cite{hu2021lora}, which can be expressed as follows:

\begin{equation}
\begin{aligned}
\mathbf{h} = \mathbf{W}_0\mathbf{x} + \Delta \mathbf{W} \mathbf{x}
\end{aligned}
\label{eq:lora}
\end{equation}

According to Eq. \ref{eq:TF}, we use LoRA to fine-tune the Linear operation in FFN can be written as:

\begin{equation}
\begin{aligned}
\mathbf{x}_{i+1}&=(\mathbf{W}_{i+1,1}'\mathbf{x}_0+\mathbf{b}_{i+1,1})+\sum_{j=1}^{i+1}(\mathbf{W}'_{j,3}+\Delta\mathbf{W}_{j+1,3})\\
&+\sigma ((\mathbf{W}'_{j,2}+\Delta\mathbf{W}_{j+1,2})\mathbf{x}'_{0}+\mathbf{b}'_{j,2})
\end{aligned}
\label{eq:lora-UAT}
\end{equation}

From Eq. \ref{eq:lora-UAT}, it can be seen that LoRA essentially fine-tunes the DUAT parameters layer by layer for a specific task.

\section{Conclusion}
\label{section: Conclusion}
In this paper, we delve into the theoretical underpinnings of LLMs, demonstrating that contemporary LLMs, primarily constructed with Transformer architectures, embody concrete manifestations of the DUAT. The remarkable generalization prowess exhibited by LLMs is attributed to their MHA modules and residual operation, which enable the adaptation to approximate diverse functions based on the presented input data. Contextual interaction emerges as a paramount capability for LLMs, manifesting such abilities as ICL, instruction following, and contextual reasoning. These competencies are enabled by the Transformer's MHA to learn from context.

Expanding upon this understanding, we have provided a rigorous theoretical grounding for key techniques employed in LLMs, including LoRA for efficient fine-tuning and pruning for model compression, elucidating their effectiveness through the lens of the UAT. By leveraging the theoretical framework provided by the DUAT, not only can existing methodologies be explained but also avenues for the future evolution of LLMs are illuminated.

\section{Rethinking LLMs}\label{section:Rethinking LLMs and future study}

\begin{figure}[htbp!]
\centering
\includegraphics[width=0.65\textwidth]{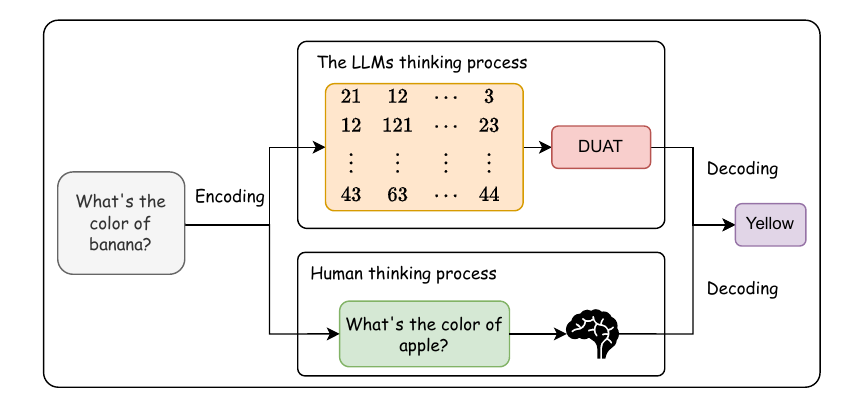}
\caption{The comparison between Human and LLMs.} 
\label{fig:HumanVSLLMs}
\vspace{-1.5em}
\end{figure}

The capabilities of LLMs have become so advanced that their language processing abilities are approaching human levels, raising a core question: How do LLMs differ from humans in language processing? Figure \ref{fig:HumanVSLLMs} illustrates the comparison between the language processing processes of LLMs and humans. Both start with language encoding—humans encode language through a character-based system, while LLMs use numerical arrays. At this level, there is almost no difference. Given the ambiguity of words, determining context is crucial: humans understand context through the activation and transmission of neurons in the brain, while LLMs approximate the corresponding functions using DUAT. Here, the input and output of network layers are analogous to the transmission of neural signals in the brain, and the function approximation corresponds to the final output of humans. From this perspective, the differences between humans and LLMs in language processing seem minimal.

So how do we explain human understanding, analysis of language, and memory retrieval? Are these also capabilities of LLMs? First, we can consider the human brain as a combination of one or more DUAT models, which are randomly initialized at birth. What supports such an assumption? The original neural network, the perceptron \cite{Rosenblatt1963PRINCIPLESON}, was designed based on the human neuron. Over time, as deep learning networks evolved, the difference in form between deep networks and the perceptron grew, leading people to no longer associate neural networks with human neurons. However, through our derivation, the Transformer can also be understood as an implementation of DUAT, and the perceptron can be seen as a simple UAT. So the human brain can be considered as a cluster of multi-layer perception. Learning language in a social environment is akin to training the DUATs in the brain. Therefore, what we call understanding, analysis, and memory retrieval are essentially processes of fitting outputs based on inputs. Taking memory retrieval as an example: This is essentially fitting specific results based on particular words, as there is no actual database in the brain. For instance, recalling the experience of eating an apple for the first time in childhood is the brain fitting specific results based on those words; without those words, the memory would not surface. This is because the brain dynamically outputs results based on inputs and fills in the details of the event based on learned natural rules (We provide some examples about understanding, analysis of language, and memory retrieval in \textcolor{blue}{Appendix E}). However, as humans grow, the weights in the brain are constantly updated, leading to potential memory distortions \citep{Wang2024SchrodingersML}.

Therefore, it is entirely reasonable for LLMs to make errors or generate hallucinations—these are just outputs produced based on existing weights and inputs, a problem humans also face. We believe the greatest advantage humans have over LLMs is their powerful multimodal and multitask processing abilities (which can be understood as the coordination of multiple DUAT models to produce reasonable results). Another advantage is the interaction with the real world, which allows us to verify the knowledge we've learned in reality, enabling the brain to continuously optimize its parameters based on inputs. LLMs, on the other hand, are limited to function approximation within the corpora data.

%

\clearpage
\bibliography{iclr2025_conference}
\bibliographystyle{iclr2025_conference}

\clearpage
\appendix
\setcounter{figure}{0}
\setcounter{table}{0}

\section{The Matrix-Vector Form of MHA}
\label{section:The Sparsity}
In this section, we will present the transformation process of MHA into their corresponding matrix-vector forms.

\begin{figure*}[p]
\centering
\includegraphics[width=0.9\textwidth]{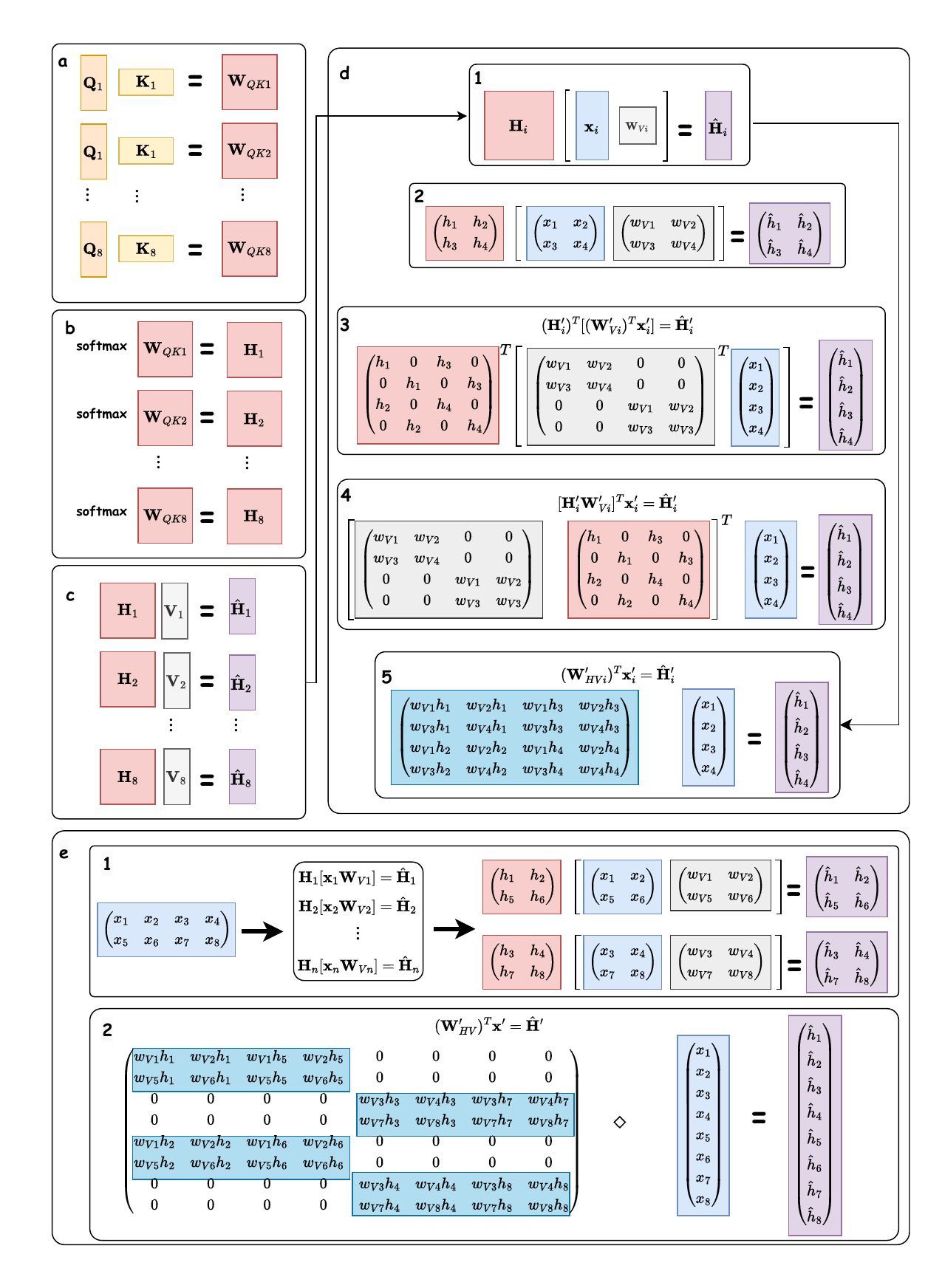}
\caption{The process of transforming $\text{Concat}(\hat{\mathbf{H}}_1...\hat{\mathbf{H}}_8)$ in the MHA into its corresponding matrix-vector form $\mathbf{W}_{HV}'\mathbf{x}'=\hat{\mathbf{H}}'$.} 
\label{fig:TF2M}
\vspace{-1em}
\end{figure*}

Figure \ref{fig:TF2M}.\textbf{a} represents the computation process of obtaining $\mathbf{W}_{QKi}$ in the attention mechanism, while Figure \ref{fig:TF2M}.\textbf{b} computes $\mathbf{H}_{1}, \dots, \mathbf{H}_{8}$ through the softmax operation. Figure \ref{fig:TF2M}.\textbf{c} and Figure \ref{fig:TF2M}.\textbf{e} illustrates the process of converting $\operatorname{Concat}\left(\mathbf{H}_1\mathbf{V}_1, \ldots, \mathbf{H}_{h}\mathbf{V}_h\right)$ into the matrix-vector form $(\mathbf{W}_{HV}')^T\mathbf{x}'=\hat{\mathbf{H}}'$, where $\mathbf{W}_{HV}'$ is generated from $\mathbf{H}_{1}, \dots, \mathbf{H}_{8}$ and $\mathbf{W}_{V1}, \dots, \mathbf{W}_{V8}$. Figure \ref{fig:TF2M}.\textbf{c} shows $\mathbf{H}_i[\mathbf{x}_i\mathbf{W}_{Vi}]$. In Figure \ref{fig:TF2M}.\textbf{d}, we provide a simple example demonstrating the conversion of $\mathbf{H}_i[\mathbf{x}_i\mathbf{W}_{Vi}]$ into $(\mathbf{W}_{HVi}')^T\mathbf{x}_i'$. This process is divided into four parts: Figure \ref{fig:TF2M}.\textbf{d.1} represents the general form of $\mathbf{H}_i[\mathbf{x}_i\mathbf{W}_{Vi}]$, while Figure \ref{fig:TF2M}.\textbf{d.2} serves as a simple example of Figure \ref{fig:TF2M}.\textbf{d.1}. Figure \ref{fig:TF2M}.\textbf{d.3} first rewrites Figure \ref{fig:TF2M}.\textbf{d.2} using matrix multiplication as $[\mathbf{H}_i'\mathbf{W}_{Vi}')]^T \mathbf{x}_i'=\hat{\mathbf{H}}_i'$, and then express it as $(\mathbf{H}_i')^T[\mathbf{W}_{Vi}')^T \mathbf{x}_i']=\hat{\mathbf{H}}_i'$ in Figure \ref{fig:TF2M}.\textbf{d.4}. Finally, Figure \ref{fig:TF2M}.\textbf{d.5} is obtained as $(\mathbf{W}_{HVi}')^T\mathbf{x}_i'=(\hat{\mathbf{H}}_i')^T $. It can be observed that $\mathbf{W}_{HVi}'$ is a dense matrix. In Figure \ref{fig:TF2M}.\textbf{e}, we present a simple example of $\mathbf{H}_1[\mathbf{x}_1\mathbf{W}_{V1}], \mathbf{H}_2[\mathbf{x}_2\mathbf{W}_{V2}]...$ into the matrix-vector form $(\mathbf{W}_{HV}')^T\mathbf{x}'=(\hat{\mathbf{H}})^T $. Figure \ref{fig:TF2M}.\textbf{e.1} depicts $\mathbf{H}_i[\mathbf{x}_i\mathbf{W}_{Vi}]$, while Figure \ref{fig:TF2M}.\textbf{e.2} represents the Matrix-vector format of Figure \ref{fig:TF2M}.\textbf{e.1}. 

\section{Transformer: The implementation of DUAT}
In this section, we will prove multi-layer Transformer is the implementation of DUAT.

\subsection{The Properties of UAT}
\label{section:The Properties of UAT}
Before expressing residual-based CNNs and Transformer-based ViTs in the UAT format, we present a lemma regarding UAT. There are two cases for UAT-approximated functions: $ f(\mathbf{x}) \in \mathbb{R} $ and $ f(\mathbf{x}) \in \mathbb{R}^m $. The proof for the case where $ f(\mathbf{x}) \in \mathbb{R} $ can be inferred from $ f(\mathbf{x}) \in \mathbb{R}^m $. Therefore, we will only provide the proof for approximating $ f(\mathbf{x}) \in \mathbb{R}^m $ using UAT.

Lemma 1. The mathematical form of UAT remains unchanged when multiplied by a matrix (constant).

\begin{equation}
\begin{aligned}
G(\mathbf{x}) &= \mathbf{\beta} \sum_{j  = 1}^N \mathbf{\alpha}_j \sigma\left(\mathbf{W}_j^{\mathrm{T}} \mathbf{x}+\mathbf{b}_j\right)\\
&= \sum_{j  = 1}^N \mathbf{\beta}\mathbf{\alpha}_j \sigma\left(\mathbf{W}_j^{\mathrm{T}} \mathbf{x}+\mathbf{b}_j\right)\\
\end{aligned}
\label{eq:UAT*W}
\end{equation}

Eq. \ref{eq:UAT*W} shows the representation of UAT multiplying a matrix. Let $ \mathbf{\alpha}_j = \mathbf{\beta}\mathbf{\alpha}_j $, and the general mathematical form of Eq. \ref{eq:UAT*W} remains consistent with the original UAT mathematical form. Thus, it is proven that the mathematical form of UAT remains unchanged when multiplied by a matrix (constant).

\subsection{From UAT to DUAT}
\label{section:From UAT to DUAT}
In this section, we first present a general form for a single-layer residual term in a network and demonstrate that the mathematical form of a multi-layer network composed of this residual term aligns with the DUAT framework. Before proceeding with the formal proof, we first define DUAT explicitly: DUAT shares the overall mathematical form of UAT, but with certain parameters influenced by the input, allowing them to change dynamically in response to it. We refer to these parameters as "dynamic parameters." In standard UAT, parameters remain fixed once training is complete, whereas, in DUAT, the dynamic parameters are functions of the input and thus vary with it. Consequently, dynamic parameters in DUAT may take the form of complex functions. Currently, DUAT primarily is implemented by residual structure, meaning that, in general, these complex functions are also DUAT. Next, we will give the proof.

A general residual term of the network can be written as: 

\begin{equation}
\begin{aligned}
\mathbf{x}'_{i} = (\mathbf{W}'_{i,1}\mathbf{x}'_{i-1} + \mathbf{b}'_{i,3}) + \mathbf{W}'_{i,3}\sigma (\mathbf{W}'_{i,2}\mathbf{x}'_{i-1} + \mathbf{b}'_{i,2})
\end{aligned}
\label{eq:res-term}
\end{equation}

where $i = 1, 2, 3, \ldots $ and $\mathbf{x}'_{0}$ represents the input. A multi-layer network structured in this way aligns with the mathematical form of DUAT. To demonstrate that a multi-layer network leverages the general term of Eq. \ref{eq:res-term} corresponds to DUAT, we first examine the forms of single-layer and two-layer networks, as shown in Eqs. \ref{eq:R-app-1} and \ref{eq:R-app-2}.

\begin{equation}
\centering
\begin{aligned}
\mathbf{x}'_{1} =(\mathbf{W}'_{1,1}\mathbf{x}'_{0}+\mathbf{b}'_{1,3})+\mathbf{W}'_{1,3}\sigma (\mathbf{W}'_{1,2}\mathbf{x}'_{0}+\mathbf{b}'_{1,2})
\end{aligned}
\label{eq:R-app-1}
\end{equation}

\begin{figure*}[htbp!]
\centering
\begin{equation}
\centering
\begin{aligned}
\mathbf{x}'_{2} & =\mathbf{W}'_{2,1}\mathbf{x}'_{1}+\mathbf{W}'_{2,3}\sigma (\mathbf{W}'_{2,2}\mathbf{x}'_{1}+\mathbf{b}'_{2,2})+\mathbf{b}'_{2,3}\\
&=(\mathbf{W}'_{2,1}\mathbf{x}'_{1}+\mathbf{b}'_{2,3})+\mathbf{W}'_{2,3}\sigma (\mathbf{W}'_{2,2}\mathbf{x}'_{1}+\mathbf{b}'_{2,2})\\
&=\{\mathbf{W}'_{2,1}[(\mathbf{W}'_{1,1}\mathbf{x}'_{0}+\mathbf{b}'_{1,3})+\mathbf{W}'_{1,3}\sigma (\mathbf{W}'_{1,2}\mathbf{x}'_{0}+\mathbf{b}'_{1,2})]+\mathbf{b}'_{2,3}\}\\
&+\mathbf{W}'_{2,3}\sigma \{\mathbf{W}'_{2,2}[(\mathbf{W}'_{1,1}\mathbf{x}'_{0}+\mathbf{b}'_{1,3})+\mathbf{W}'_{1,3}\sigma (\mathbf{W}'_{1,2}\mathbf{x}'_{0}+\mathbf{b}'_{1,2})]+\mathbf{b}'_{2,2}\}\\
&=\{\mathbf{W}'_{2,1}(\mathbf{W}'_{1,1}\mathbf{x}'_{0}+\mathbf{b}'_{1,3})+\mathbf{b}'_{2,3}+\mathbf{W}'_{2,1}\mathbf{W}'_{1,3}\sigma (\mathbf{W}'_{1,2}\mathbf{x}'_{0}+\mathbf{b}'_{1,2})\}\\
&+\mathbf{W}'_{2,3}\sigma \{\mathbf{W}'_{2,2}({\mathbf{W}}'_{1,1}\mathbf{x}'_{0}+\mathbf{b}'_{1,3})+\mathbf{b}'_{2,2}+\mathbf{W}'_{2,2}\mathbf{W}'_{1,3}\sigma (\mathbf{W}'_{1,2}\mathbf{x}'_{0}+\mathbf{b}'_{1,2})\}\\
&=\{(\underline{\mathbf{W}'_{2,1}\mathbf{W}'_{1,1}}\mathbf{x}'_{0}+\underline{\mathbf{W}'_{2,1}\mathbf{b}'_{1,3}+\mathbf{b}'_{2,3}})+\underline{\mathbf{W}'_{2,1}\mathbf{W}'_{1,3}}\sigma (\mathbf{W}'_{1,2}\mathbf{x}'_{0}+\mathbf{b}'_{1,2})\}\\
&+\mathbf{W}'_{2,3}\sigma \{\underline{\mathbf{W}'_{2,2}\mathbf{W}'_{1,1}}\mathbf{x}'_{0}+\underline{(\mathbf{W}'_{2,2}\mathbf{b}'_{1,3}+\mathbf{b}'_{2,2})+\mathbf{W}'_{2,2}\mathbf{W}'_{1,3}\sigma (\mathbf{W}'_{1,2}\mathbf{x}'_{0}+\mathbf{b}'_{1,2})}\}\\
\end{aligned}
\label{eq:R-app-2}
\end{equation}
\end{figure*}

In Eq. \ref{eq:R-app-2}, let $ \mathbf{W}'_{2,1}=\mathbf{W}'_{2,1}\mathbf{W}'_{1,1} $,
$ \mathbf{b}'_{2,1}=\mathbf{W}'_{2,1}\mathbf{b}'_{1,3}+\mathbf{b}'_{2,3} $,
$ \mathbf{W}'_{1,3}=\mathbf{W}'_{2,1}\mathbf{W}'_{1,3} $,
$ \mathbf{W}'_{2,2}=\mathbf{W}'_{2,2}\mathbf{W}'_{1,1} $, and
$ \mathbf{b}'_{2,2}=(\mathbf{W}'_{2,2}\mathbf{b}'_{1,3}+\mathbf{b}'_{2,2})+\mathbf{W}'_{2,2}\mathbf{W}'_{1,3}\sigma (\mathbf{W}'_{1,2}\mathbf{x}'_{0}+\mathbf{b}'_{1,2}) $. Thus, Eq. \ref{eq:R-app-2} can be written into Eq. \ref{eq:UAT-2-layer}.

\begin{equation}
\begin{aligned}
\mathbf{x}'_{2}&=(\mathbf{W}'_{2,1}\mathbf{x}'_{0}+\mathbf{b}'_{2,1})+\mathbf{W}'_{1,3}\sigma (\mathbf{W}'_{1,2}\mathbf{x}'_{0}+\mathbf{b}'_{1,2})\\
&+\mathbf{W}'_{2,3}\sigma (\mathbf{W}'_{2,2}\mathbf{x}'_{0}+\mathbf{b}'_{2,2})
\end{aligned}
\label{eq:UAT-2-layer}
\end{equation}

For the cleatity, we could write Eq. \ref{eq:R-app-1} and Eq. \ref{eq:UAT-2-layer} into Eq. \ref{eq:UAT-1-layer_R} and Eq. \ref{eq:UAT-2-layer_R}, where $UAT^R_1=\mathbf{W}'_{1,3}\sigma (\mathbf{W}'_{1,2}\mathbf{x}'_{0}+\mathbf{b}'_{1,2})$ and $UAT^R_2=\Sigma_{j=1}^2\mathbf{W}'_{j,3}\sigma (\mathbf{W}'_{j,2}\mathbf{x}'_{0}+\mathbf{b}'_{j,2})
$.

\begin{equation}
\begin{aligned}
\mathbf{x}'_{1} =(\mathbf{W}'_{1,1}\mathbf{x}'_{0}+\mathbf{b}'_{1,3})+UAT^R_1
\end{aligned}
\label{eq:UAT-1-layer_R}
\end{equation}

\begin{equation}
\begin{aligned}
\mathbf{x}'_{2}&=(\mathbf{W}'_{2,1}\mathbf{x}'_{0}+\mathbf{b}'_{2,1})+UAT^R_2
\end{aligned}
\label{eq:UAT-2-layer_R}
\end{equation}

According to Eq. \ref{eq:UAT-1-layer_R} and Eq. \ref{eq:UAT-2-layer_R}, aside from the initial term, the overall mathematical forms of the residual terms $ UAT^R_1 $ and $ UAT^R_2 $ are consistent with UAT. However, for $ UAT^R_2 $, its parameter $ \mathbf{b}'_{2,2} $ is influenced by the input, while other weight parameters, such as $ \mathbf{W}'_{2,1} $, can be disregarded. These parameters are only affected by other parameters, so once the network training is complete, they are essentially fixed. This allows us to focus primarily on the dynamic parameters. In equation $ \mathbf{b}'_{2,2} = (\mathbf{W}'_{2,2}\mathbf{b}'_{1,3} + \mathbf{b}'_{2,2}) + \mathbf{W}'_{2,2}\mathbf{W}'_{1,3} \sigma (\mathbf{W}'_{1,2}\mathbf{x}'_{0} + \mathbf{b}'_{1,2}) $, except for the initial term, $ \mathbf{b}'_{2,2} $ also follows the same mathematical form as UAT. This can be interpreted as dynamically adjusting the bias term $ \mathbf{b}'_{2,2} $ via UAT based on the input. In conclusion, we have demonstrated that the mathematical forms of one-layer and two-layer residual networks are consistent with the UAT framework and the two-layer residual network is the DUAT function.

Next, we will use mathematical induction to prove that the mathematical form of a multi-layer residual network is also a DUAT function. Assume that the overall mathematical form of the first $ i $ layers of the residual network aligns with UAT. Our goal is to show that the overall mathematical form of the $ i+1 $-th layer remains consistent with UAT.

For clarity, we make the following definitions: since the overall mathematical form of the first $ i $ layers is consistent with UAT, we can write it as $ \mathbf{x}'_{i} = (\mathbf{W}'_{i,1} \mathbf{x}'_{0} + \mathbf{b}'_{i,1}) + UAT^R_i $, where the first term is explicitly written, and the remainder is denoted as $ UAT^R_i $, with $ UAT_{i}^R = \sum_{j=1}^{i} \mathbf{W}'_{j,2} \sigma (\mathbf{W}'_{j,1} \mathbf{x}'_{0} + \mathbf{b}'_{j,1}) $. Since $ \mathbf{x}'_{i+1} = (\mathbf{W}'_{i+1,1} \mathbf{x}'_{i} + \mathbf{b}'_{i+1,3}) + \mathbf{W}'_{i+1,3} \sigma (\mathbf{W}'_{i+1,2} \mathbf{x}'_{i} + \mathbf{b}'_{i+1,2}) $, we divide $ \mathbf{x}'_{i+1} $ into two parts: $ (\mathbf{W}'_{i+1,1} \mathbf{x}'_{i} + \mathbf{b}'_{i+1,3}) $ and $ \mathbf{W}'_{i+1,3} \sigma (\mathbf{W}'_{i+1,2} \mathbf{x}'_{i} + \mathbf{b}'_{i+1,2}) $.

First, consider $ (\mathbf{W}'_{i+1,1} \mathbf{x}'_{i} + \mathbf{b}'_{i+1,3}) $. Substituting $ \mathbf{x}'_{i} = (\mathbf{W}'_{i,1} \mathbf{x}'_{0} + \mathbf{b}'_{i,1}) + UAT^R_i $, we obtain Eq. \ref{eq:i+1_term_1}. Setting $ \mathbf{W}'_{i+1,1} = \mathbf{W}'_{i+1,1} \mathbf{W}'_{i,1} $ and $ \mathbf{b}'_{i+1,1} = \mathbf{W}'_{i+1,1} \mathbf{b}'_{i,1} + \mathbf{b}'_{i+1,3} $, we can simplify the first part to $ (\mathbf{W}'_{i+1,1} \mathbf{x}'_{0} + \mathbf{b}'_{i+1,1}) + \mathbf{W}'_{i+1,1} UAT^R_i $. Since the overall mathematical form of $ UAT^R_i $ is consistent with the $ i $-layer UAT and we have shown that the UAT form is preserved when multiplied by a matrix, we have thus demonstrated that the overall mathematical form of the first part remains consistent with UAT.

\begin{figure*}[htbp!]
\centering
\begin{equation}
\begin{aligned}
&(\mathbf{W}'_{i+1,1}\mathbf{x}'_{i} + \mathbf{b}'_{i+1,3})\\
=&\{\mathbf{W}'_{i+1,1}[(\mathbf{W}'_{i,1}\mathbf{x}'_{0}+\mathbf{b}'_{i,1})+UAT^R_i] + \mathbf{b}'_{i+1,3}\}\\
=&\{\mathbf{W}'_{i+1,1}(\mathbf{W}'_{i,1}\mathbf{x}'_{0}+\mathbf{b}'_{i,1})+\mathbf{W}'_{i+1,1}UAT^R_i + \mathbf{b}'_{i+1,3}\}\\
=&\{[\underline{ \mathbf{W}'_{i+1,1}\mathbf{W}'_{i,1}}\mathbf{x}'_{0}+\underline{(\mathbf{W}'_{i+1,1}\mathbf{b}'_{i,1}+ \mathbf{b}'_{i+1,3})}]+\mathbf{W}'_{i+1,1}UAT^R_i \}\\
\end{aligned}
\label{eq:i+1_term_1}
\end{equation}
\end{figure*}

Next, we show that the overall mathematical form of the second part, $ \mathbf{W}'_{i+1,3} \sigma (\mathbf{W}'_{i+1,2} \mathbf{x}'_{i} + \mathbf{b}'_{i+1,2}) $, is also consistent with UAT. Substituting $ \mathbf{x}'_{i} = (\mathbf{W}'_{i,1} \mathbf{x}'_{0} + \mathbf{b}'_{i,1}) + UAT^R_i $ into this expression, we arrive at Eq. \ref{eq:i+1_term_2}. Setting $ \mathbf{W}'_{i+1,2} = \mathbf{W}'_{i+1,2} \mathbf{W}'_{i,1} $ and $ \mathbf{b}'_{i+1,2} = (\mathbf{W}'_{i+1,2} \mathbf{b}'_{i,1} + \mathbf{b}'_{i+1,2}) + \mathbf{W}'_{i+1,2} UAT^R_i $, we can rewrite the second part as $ \mathbf{W}'_{i+1,3} \sigma (\mathbf{W}'_{i+1,2} \mathbf{x}'_{0} + \mathbf{b}'_{i+1,2}) $, which can be interpreted as a term in UAT. Here, $ \mathbf{b}'_{i+1,2} $ serves as a bias term approximated using DUAT.

\begin{figure*}[htbp!]
\centering
\begin{equation}
\begin{aligned}
&\mathbf{W}'_{i+1,3}\sigma (\mathbf{W}'_{i+1,2}\mathbf{x}'_{i} + \mathbf{b}'_{i+1,2})\\
=&\mathbf{W}'_{i+1,3}\sigma \{\mathbf{W}'_{i+1,2}[(\mathbf{W}'_{i,1}\mathbf{x}'_{0}+\mathbf{b}'_{i,1})+UAT^R_i] + \mathbf{b}'_{i+1,2}\}\\
=&\mathbf{W}'_{i+1,3}\sigma \{\mathbf{W}'_{i+1,2}(\mathbf{W}'_{i,1}\mathbf{x}'_{0}+\mathbf{b}'_{i,1})
+\mathbf{W}'_{i+1,2}UAT^R_i + \mathbf{b}'_{i+1,2}\}\\
=&\mathbf{W}'_{i+1,3}\sigma \{\underline{\mathbf{W}'_{i+1,2}\mathbf{W}'_{i,1}}\mathbf{x}'_{0}+\underline{(\mathbf{W}'_{i+1,2}\mathbf{b}'_{i,1}+ \mathbf{b}'_{i+1,2})+\mathbf{W}'_{i+1,2}UAT^R_i} \}\\
\end{aligned}
\label{eq:i+1_term_2}
\end{equation}
\end{figure*}

Let $ UAT^R_{i+1} = \mathbf{W}'_{i+1,1} UAT^R_i + \mathbf{W}'_{i+1,3} \sigma (\mathbf{W}'_{i+1,2} \mathbf{x}'_{0} + \mathbf{b}'_{i+1,2}) $. Then, we can express $\mathbf{x}'_{i+1}$ as $ \mathbf{x}'_{i+1} = (\mathbf{W}'_{i+1,1} \mathbf{x}'_{0} + \mathbf{b}'_{i+1,1}) + UAT^R_{i+1}$. Thus, apart from the initial term, the overall mathematical form of $\mathbf{x}'_{i+1}$ remains consistent with UAT. Since a single term does not alter the nature of UAT, we conclude that the overall form of a multi-layer residual network aligns with UAT. Furthermore, as certain bias parameters within the network, such as $\mathbf{b}'_{i+1,2}$, are also approximated using UAT, we have therefore demonstrated that the mathematical form of a multi-layer residual network is a UAT function.

\subsection{Transformer to DUAT}
\label{section:Transformer to UAT}

Similarly, to prove that Transformer-based ViTs are also DUAT functions, we start by deriving their general form based on Figure 9 in the main text. Transformers involve two key operations: MHA and the Feed-Forward Network (FFN). In matrix-vector form, these operations correspond to Equations \ref{eq:MHA} and \ref{eq:FFN}.

\begin{equation}
MHA(\mathbf{x}_{i}) \mapsto \mathbf{W}'_{i,1}\mathbf{x}'_{i}
\label{eq:MHA}
\end{equation}

\begin{equation}
FFN(\mathbf{x}_{i}) \mapsto \mathbf{W}'_{i,3}\sigma (\mathbf{W}'_{i,2}\mathbf{x}'_{i}+\mathbf{b}'_{i,2})+\mathbf{b}'_{i,3}\\
\label{eq:FFN}
\end{equation}

\begin{figure*}[htbp!]
\centering
\begin{equation}
\centering
\begin{aligned}
\mathbf{x}'_{i+1} & =\mathbf{W}'_{i+1,1}\mathbf{x}'_{i}+\mathbf{W}'_{i+1,3}\sigma [\mathbf{W}'_{i+1,2}(\mathbf{W}'_{i+1,1}\mathbf{x}'_{i})+\mathbf{b}'_{i+1,2}]+\mathbf{b}'_{i+1,3}\\
&=\mathbf{W}'_{i+1,1}\mathbf{x}'_{i}+\mathbf{W}'_{i+1,3}\sigma (\underline{ \mathbf{W}'_{i+1,2}\mathbf{W}'_{i+1,1}}\mathbf{x}'_{i}+\mathbf{b}'_{i+1,2})+\mathbf{b}'_{i+1,3}\\
\end{aligned}
\label{eq:TF-general-term}
\end{equation}
\end{figure*}

Thus, a general term in a Transformer-based network can be expressed as Equation \ref{eq:TF-general-term}. Letting $\mathbf{W}'_{i+1,2} = \mathbf{W}'_{i+1,2} \mathbf{W}'_{i+1,1}$, we can rewrite the general term as follows:

\begin{equation}
\mathbf{x}'_{i+1} = (\mathbf{W}'_{i+1,1} \mathbf{x}'_{i} + \mathbf{b}'_{i+1,3}) + \mathbf{W}'_{i,3} \sigma (\mathbf{W}'_{i+1,2} \mathbf{x}'_{i} + \mathbf{b}'_{i+1,2})
\label{eq:TF-general-term-simplify}
\end{equation}

Clearly, Eq. \ref{eq:TF-general-term-simplify} matches the mathematical form presented in Section \ref{section:From UAT to DUAT}: $\mathbf{x}'_{i+1} = (\mathbf{W}'_{i+1,1} \mathbf{x}'_{i} + \mathbf{b}'_{i+1,3}) + \mathbf{W}'_{i+1,3} \sigma (\mathbf{W}'_{i+1,2} \mathbf{x}'_{i} + \mathbf{b}'_{i+1,2})$. Therefore, a multilayer Transformer-based ViT is indeed DUAT function.

To clearly show the DUAT format of a multilayer ViT, we use the approach outlined in Section \ref{section:From UAT to DUAT}. Assuming that the overall mathematical form of $\mathbf{x}'_i$ aligns with the UAT structure, we decompose $\mathbf{x}'_i$ into a main term and a residual term, expressed as $\mathbf{x}'_i = (\mathbf{W}_{i,1} \mathbf{x}_0 + \mathbf{b}_{i,3}) + UAT^R_i$, where $UAT_{i}^R = \sum_{j=1}^{i} \mathbf{W}'_{j,3} \sigma (\mathbf{W}'_{j,2} \mathbf{x}'_{0} + \mathbf{b}'_{j,2})$ and $i=1,2,\ldots$. Substituting this into $\mathbf{x}'_{i+1} = (\mathbf{W}'_{i+1,1} \mathbf{x}'_{i} + \mathbf{b}'_{i+1,3}) + \mathbf{W}'_{i,3} \sigma (\mathbf{W}'_{i+1,2} \mathbf{x}'_{i} + \mathbf{b}'_{i+1,2})$, we get Eq. \ref{eq:UAT-TF}.

\begin{figure*}[htbp!]
\centering
\begin{equation}
\begin{aligned}
\mathbf{x}'_{i+1}&= (\mathbf{W}'_{i+1,1}\mathbf{x}'_{i}+\mathbf{b}'_{i+1,3})+\mathbf{W}'_{i,3}\sigma (\mathbf{W}'_{i+1,2}\mathbf{x}'_{i}+\mathbf{b}'_{i+1,2})\\
&= \{\mathbf{W}'_{i+1,1}[(\mathbf{W}_{i,1}\mathbf{x}_0+\mathbf{b}_{i,1})+UAT^R_i]+\mathbf{b}'_{i+1,3}\}\\
&+\mathbf{W}'_{i,3}\sigma \{\mathbf{W}'_{i+1,2}[(\mathbf{W}_{i,1}\mathbf{x}_0+\mathbf{b}_{i,1})+UAT^R_i]+\mathbf{b}'_{i+1,2}\}\\
&= [\underline{ \mathbf{W}'_{i+1,1}\mathbf{W}_{i,1}}\mathbf{x}_0+\underline{(\mathbf{W}'_{i+1,1}\mathbf{b}_{i,1}+\mathbf{b}'_{i+1,3})}]+\mathbf{W}'_{i+1,1}UAT^R_i]\\
&+\mathbf{W}'_{i,3}\sigma \{\underline{ \mathbf{W}'_{i+1,2}\mathbf{W}_{i,1}}\mathbf{x}_0+\underline{ (\mathbf{W}'_{i+1,2}\mathbf{b}_{i,1}+\mathbf{b}'_{i+1,2})+\mathbf{W}'_{i+1,2}UAT^R_i}\}\\
\end{aligned}
\label{eq:UAT-TF}
\end{equation}
\end{figure*}

Let $\mathbf{W}'_{i+1,1} = \mathbf{W}'_{i+1,1} \mathbf{W}_{i,1}$, $\mathbf{b}'_{i+1,3} = (\mathbf{W}'_{i+1,1} \mathbf{b}_{i,1} + \mathbf{b}'_{i+1,3})$, $UAT^R_i = \mathbf{W}'_{i+1,1} UAT^R_i$, $\mathbf{W}'_{i+1,2} = \mathbf{W}'_{i+1,2} \mathbf{W}_{i,1}$, and $\mathbf{b}'_{i+1,2} = (\mathbf{W}'_{i+1,2} \mathbf{b}_{i,1} + \mathbf{b}'_{i+1,2}) + \mathbf{W}'_{i+1,2} UAT^R_i$. Substituting these, we obtain $\mathbf{x}'_{i+1} = (\mathbf{W}'_{i+1,1} \mathbf{x}_0 + \mathbf{b}'_{i+1,3}) + UAT^R_i + \mathbf{W}'_{i,3} \sigma (\mathbf{W}'_{i+1,2} \mathbf{x}_0 + \mathbf{b}'_{i+1,2})$. Since $UAT^R_i$ overall aligns with the UAT, the overall mathematical form of the $i+1$ layers' ViT also conforms to the UAT.

So the $i+1$ layers' ViT can be expressed as:
\begin{equation}
\begin{aligned}
\mathbf{x}_{i+1}=(\mathbf{W}_{i+1,1}'\mathbf{x}_0+\mathbf{b}_{i+1,3})+\sum_{j=1}^{i+1}\mathbf{W}'_{j,3}\sigma (\mathbf{W}'_{j,2}\mathbf{x}'_{0}+\mathbf{b}'_{j,2})
\end{aligned}
\label{eq:TF-app}
\end{equation}

When $ i = 0 $, the parameters are defined as follows: $ \mathbf{W}'_{1,1} = \mathbf{W}'_{1,1} $, $ \mathbf{b}'_{1,3} = \mathbf{b}'_{1,3} $, $ \mathbf{W}'_{1,3} = \mathbf{W}'_{1,3} $, $ \mathbf{W}'_{1,2} = \mathbf{W}'_{1,2} \mathbf{W}'_{1,1} $, and $ \mathbf{b}'_{1,2} = \mathbf{b}'_{1,2} $. We define the parameters for cases where $ i \geq 1 $. For each $ j = 1, 2, \ldots, i $, the updates are as follows: $ \mathbf{W}'_{j+1,1} = \mathbf{W}'_{j+1,1} \mathbf{W}_{j,1} $, $ \mathbf{b}'_{j+1,3} = \mathbf{W}'_{j+1,1} \mathbf{b}_{j,3} + \mathbf{b}'_{j+1,3} $, $ \mathbf{W}'_{j+1,2} = \mathbf{W}'_{j+1,2} \mathbf{W}_{j,1}$, and $ \mathbf{W}'_{j,3} = \mathbf{W}'_{j+1,1} \mathbf{W}'_{j,3}$. 

Additionally, for $ j = 1, 2, \ldots, i + 1 $, the bias terms $ \mathbf{b}'_{j,2} $ are updated according to:
\begin{equation}
\begin{aligned}
\mathbf{b}'_{j,2} =& (\mathbf{W}'_{j,2} \mathbf{b}'_{j-1,3} + \mathbf{b}'_{j,2}) \\
+& \mathbf{W}'_{j,2} \sum_{k=1}^{j-1} \mathbf{W}'_{k,3} \sigma (\mathbf{W}'_{k,2} \mathbf{x}'_{0} + \mathbf{b}'_{k,2}).
\end{aligned}
\label{eq:b_j2-app}
\end{equation}

\begin{figure*}[htbp!]
\centering
\includegraphics[width=0.9\textwidth]{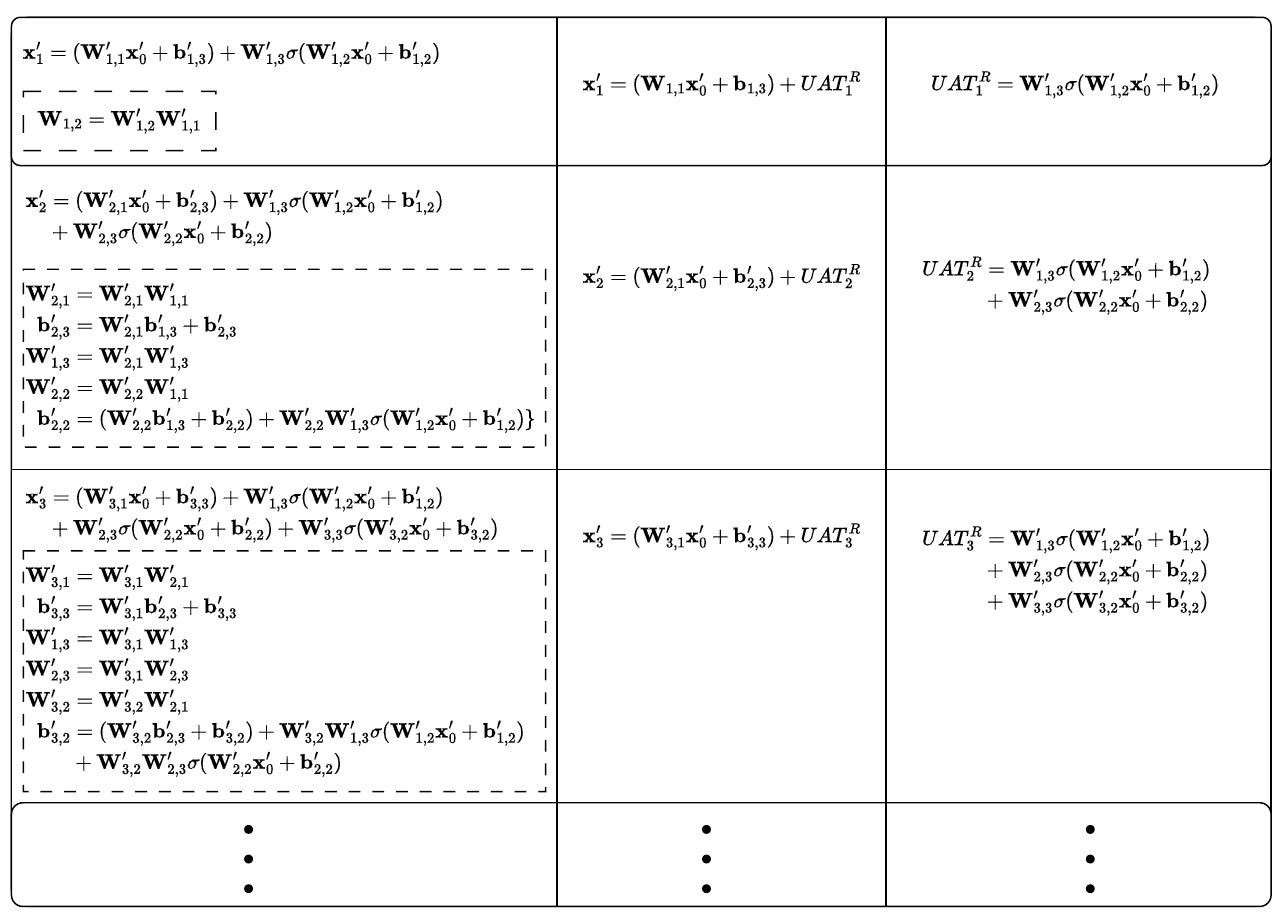}
\caption{Some examples of the DUAT format of multi-layer Transformer. The changes of parameters are represented within the dashed boxes. The parameters on the right of the equations indicate the original values, while those on the left represent the transformed values. There is no specific order of calculation for the parameters within each dashed box, but there is a top-to-bottom calculation order between different dashed boxes.
}
\label{fig:TF_UAT}
\vspace{-1em}
\end{figure*}

It means that the whole computing process is serial, if we want to compute $\mathbf{W}'_{3,1}$, we must calculate $\mathbf{W}'_{2,1}$, because $\mathbf{W}'_{3,1}=\mathbf{W}'_{3,1}\mathbf{W}'_{2,1}$. Thus, we have established that a multilayer ViT is also a DUAT function. The mathematical representation of the ViT aligns with the DUAT framework, with $\mathbf{b}'_{j,2}$ closely approximated by DUAT. The most notable difference between ViTs and residual-based CNNs is the dynamic input-dependent nature of the parameters in the MHA mechanism. Consequently, in the DUAT formulation associated with ViT, the parameters $\mathbf{W}'_{j,1}$ and $\mathbf{W}'_{j,2}$ for $i+1 \geq j \geq 1$, and $\mathbf{W}'_{j,3}$ for $i+1 > j > 1$ in layer $i$, all adapt dynamically based on the input.Besides, we provide some examples in Figure \ref{fig:TF_UAT}.


\section{The Learning Process of MHA}
\begin{figure*}[htbp]
\centering
\includegraphics[width=0.75\textwidth]{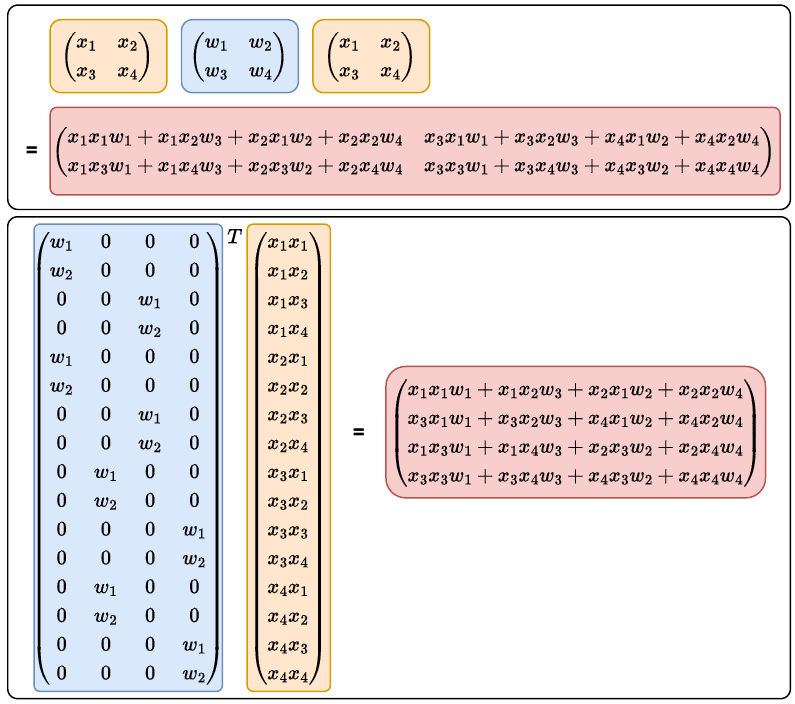}
\caption{The transformation process of the $\mathbf{H}_i$.} 
\label{fig:MVM_H_i}
\vspace{-1.5em}
\end{figure*}

In this section, we will present the specific mathematical form of the parameters $\mathbf{H}_i$ in the MHA shown in Figure \ref{fig:TF2M}. First, it can be written as Eq. \ref{eq:H_i}. Then, in Figure \ref{fig:MVM_H_i}, we illustrate how each parameter in $\mathbf{H}_i$ is calculated. It is clear that these parameters are learned elements of the covariance matrix of the input. This process disrupts the overall mathematical form of the UAT, so $\mathbf{H}_i$ can be simply and directly regarded as parameters learned from the covariance matrix of the upper-layer input.

\begin{equation}
\begin{aligned}
\mathbf{H}_i=&\operatorname{softmax}(\frac{\mathbf{x}_i \mathbf{W}_{Qi} (\mathbf{x}_i\mathbf{W}_{Ki})^T)}{\sqrt{M}}) \\
=&\operatorname{softmax}(\frac{\mathbf{x}_i \mathbf{W}_{Qi} \mathbf{W}_{Ki}^T\mathbf{x}_i^T}{\sqrt{M}}) \\
=&\operatorname{softmax}(\frac{\mathbf{x}_i \mathbf{W}_{QKi}\mathbf{x}_i^T}{\sqrt{M}}) \\
\end{aligned}
\label{eq:H_i}
\end{equation}

\begin{figure*}[htbp]
\centering
\includegraphics[width=0.95\textwidth]{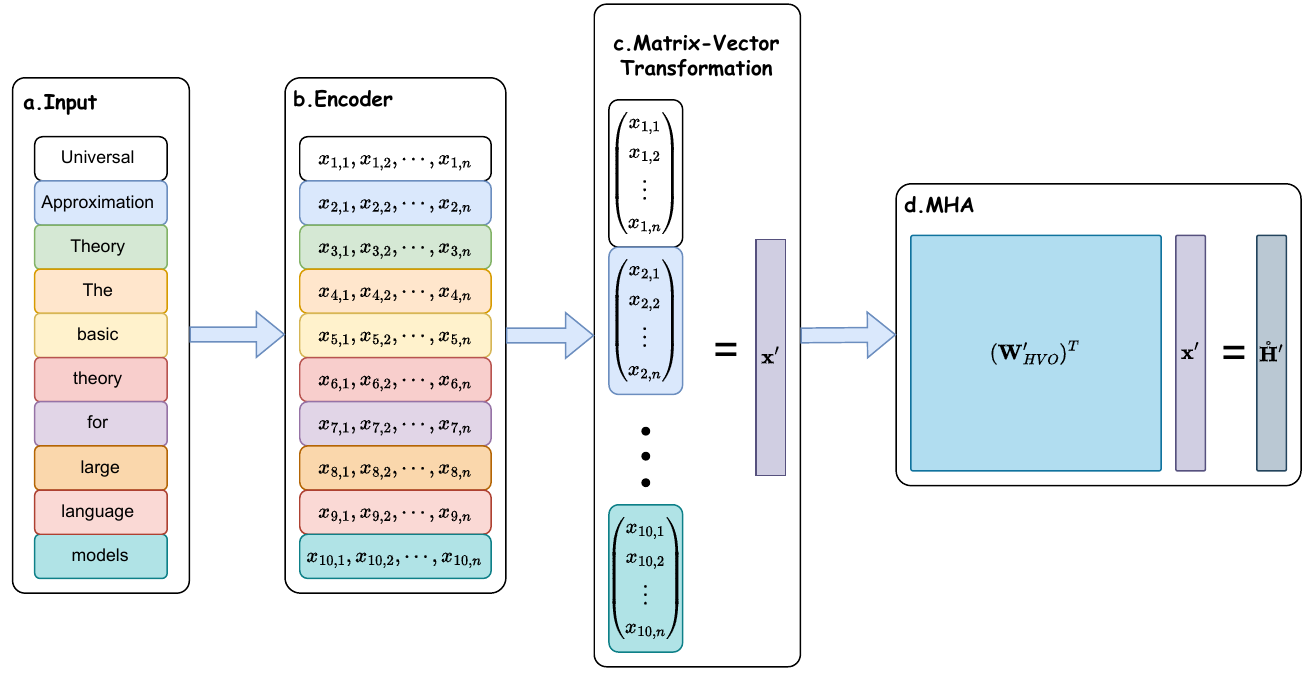}
\caption{The learning process of MHA} 
\label{fig:Text2MHA}
\vspace{-1.5em}
\end{figure*}

To more clearly illustrate the learning details of the MHA mechanism in Transformers, we present an example in Figure \ref{fig:Text2MHA}. First, the input consists of 10 characters, as shown in Figure \ref{fig:Text2MHA}.\textbf{a}. Then, in Figure \ref{fig:Text2MHA}.\textbf{b}, these characters are encoded into $n$-dimensional vectors. In Figure \ref{fig:Text2MHA}.\textbf{c}, the encoded vectors are expanded into $\mathbf{x}'$ using Matrix-Vector methods. Subsequently, in Figure \ref{fig:Text2MHA}.\textbf{d}, we compute $\mathring{\mathbf{H}}'=(\mathbf{W}_{HVO}')^T\mathbf{x}'$. Since $(\mathbf{W}_{HVO}')^T$ is an almost fully dense square matrix, each element of $\mathring{\mathbf{H}}'$ contains information from the entire input. This is why ICL exists in LLMs: each element can access the full context.

\begin{figure*}[htbp]
\centering
\includegraphics[width=0.75\textwidth]{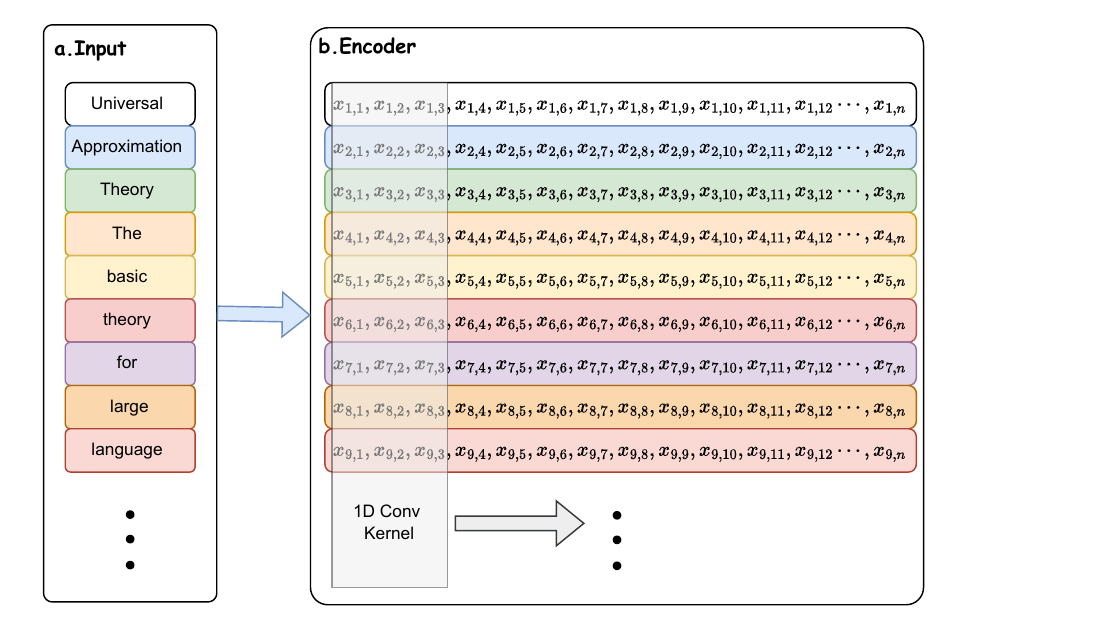}
\caption{The learning process of 1D convolution.} 
\label{fig:Conv}
\vspace{-1.5em}
\end{figure*}

While Figure \ref{fig:Conv} shows the learning process of 1D convolution. Compared to the holistic learning approach of MHA, 1D convolution learns by processing encoded text in chunks based on the size of the convolution kernel. This method is akin to learning the entire context each time but focusing on a partial set of characters of each word at a time. 

%

\section{The Comparision between Human and LLMs}

\begin{figure*}[htbp]
\centering
\includegraphics[width=0.9\textwidth]{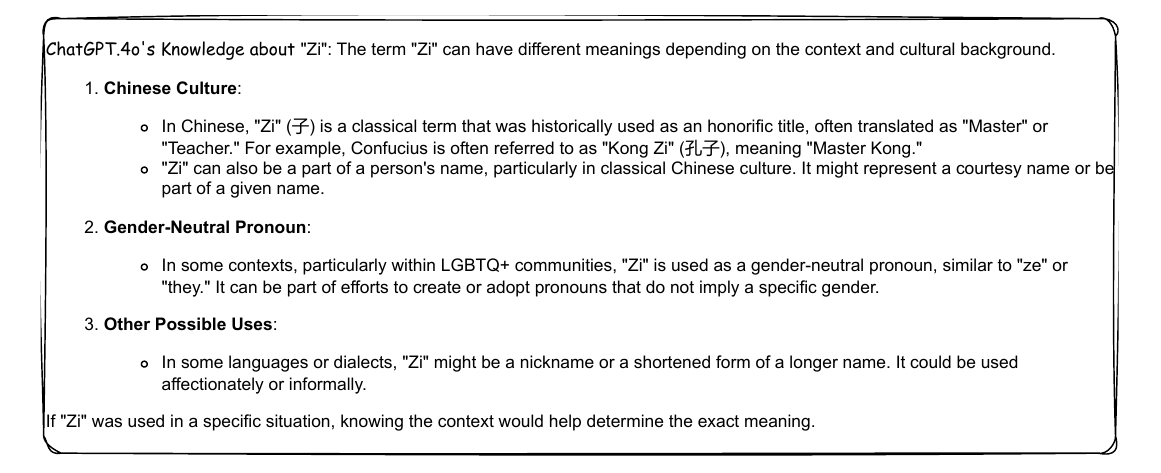}
\caption{The background information about 'zi' from Chatgpt.4o.} 
\label{fig:concept-zi}
\vspace{-1.5em}
\end{figure*}

To more specifically compare the differences between humans and LLMs, we provide a simple example here. We prepared two questions: "Nui zi created Nui zi dynamics" and "Nui zi created gravitation and relativity theory." It is important to emphasize that these sentences are entirely fictional, and there is no such person as "Nui zi." However, the term "zi" has some special meanings in specific contexts that some people might know, while others may not. We have provided some explanations of "zi" as interpreted by Chatgpt.4o in Figure \ref{fig:concept-zi}. We have offered possible answers from different groups: a kid, a general adult, a professional researcher, and also responses from Chatgpt.4o and Qwen 2.5 to the above questions.

\begin{figure*}[htbp]
\centering
\includegraphics[width=0.9\textwidth]{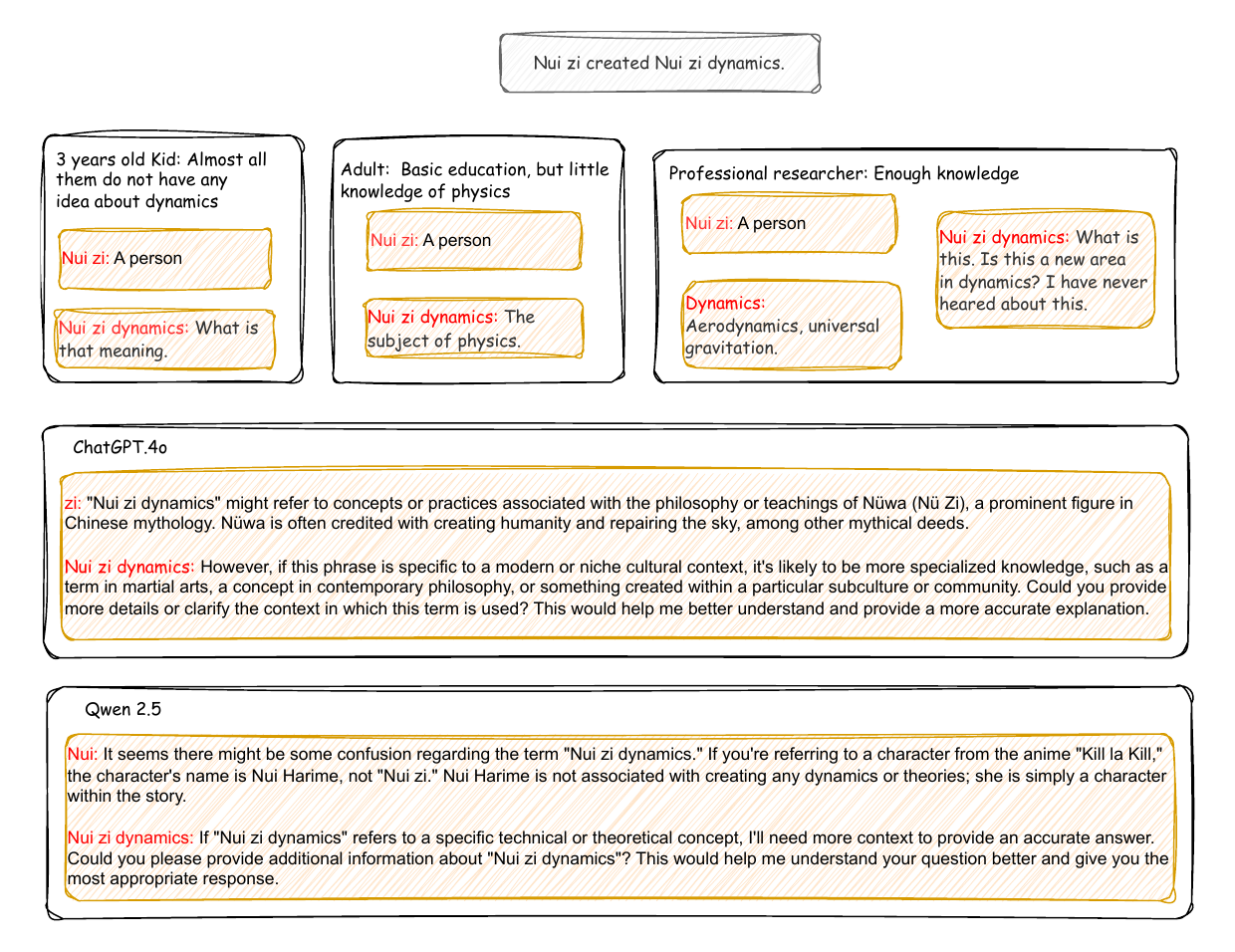}
\caption{The answers about 'Nui zi created Nui zi dynamics'.} 
\label{fig:Q1}
\vspace{-1.5em}
\end{figure*}

\begin{figure*}[htbp]
\centering
\includegraphics[width=0.9\textwidth]{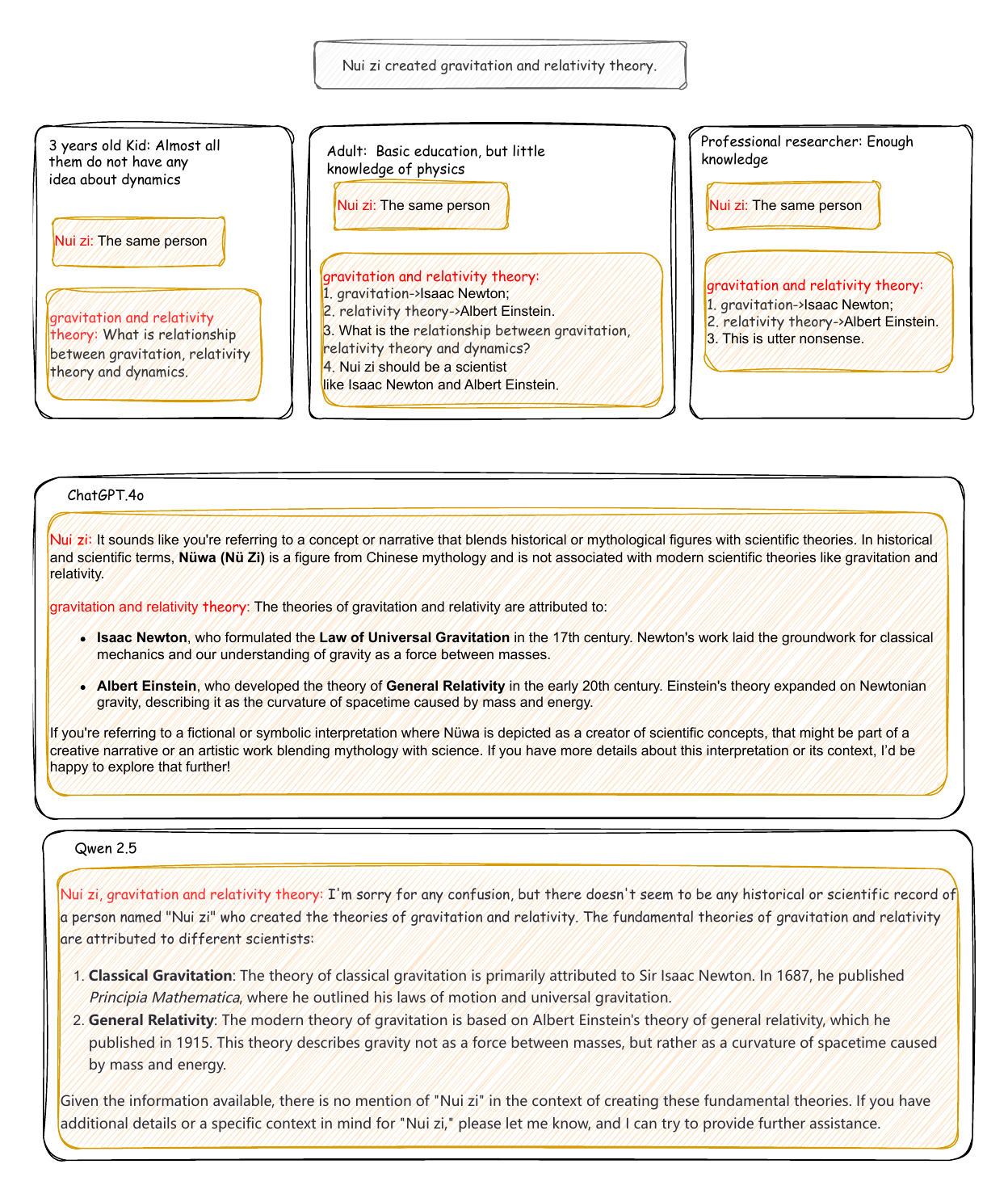}
\caption{The answers about 'Nui zi created gravitation and relativity theory'.} 
\label{fig:Q2}
\vspace{-1.5em}
\end{figure*}

Figures \ref{fig:Q1} and \ref{fig:Q2} present the answers to the two questions. Both humans and LLMs provide the output based on input. To clearly indicate which works or sentences primarily affected the generated content, we used red font to label those (this part is not generated by the model). Firstly, different individuals provide different answers based on their background knowledge, which can be understood as a process of generating output based on existing training parameters in the brain and the given input. In this process, memories about keywords lead to specific outputs, such as "Professional researcher: 1. gravitation → Isaac Newton; 2. relativity theory → Albert Einstein." This process can be understood as understanding the question, analyzing it, retrieving memories, and producing results. However, if the human brain is considered as a DUAT, it generates the output based on specific inputs, such as "gravitation → Isaac Newton" or relevant information about the theory itself. But the truth is that the brain never creates a 'memory warehouse' to store that information, it just produces the output based on input. So the 'memory' could be understood as something produced by input and pretrained parameters.

Since Chatgpt.4o and Qwen 2.5 models have been trained on extensive datasets, they possess some information about "zi," "Nui," "dynamics," "gravitation," and "relativity theory." When these characters are inputted into the models, they use their pre-trained parameters to generate relevant outputs. For the first question, they provided possible inferences based on the keywords "zi" and "Nui" and highlighted the inconsistencies in "Nui zi dynamics." For the second question, which is more detailed, both Chatgpt.4o and Qwen 2.5 were able to provide more detailed contextual information (note that the models can generate additional content based on both the input and the output from the input), which can be understood as human-like understanding, analysis, and memory retrieval.

\end{document}